\documentclass[10pt]{article}
\usepackage[margin=1in]{geometry}
\usepackage{mathpazo}
\usepackage{booktabs}
\usepackage{afterpage}

\usepackage{bm}
\usepackage{amsmath,amssymb,amsthm}
\usepackage{xspace}
\usepackage{graphicx}
\usepackage{color}
\usepackage[colorlinks=true,linkcolor=blue,citecolor=blue,urlcolor=blue]{hyperref}
\usepackage{todonotes}
\usepackage{subcaption}
\usepackage{comment}
\usepackage[ruled]{algorithm2e}
\usepackage[T1]{fontenc}
\usepackage[backend=biber,style=alphabetic,natbib=true,maxbibnames=99,minalphanames=3,maxcitenames=2]{biblatex}
\title{When does Bias Transfer in Transfer Learning?}
\author{
    Hadi Salman\thanks{Equal contribution.} \\
	MIT \\
	\texttt{hady@mit.edu} \\
	\and
	Saachi Jain\footnotemark[1] \\
	MIT \\
	\texttt{saachij@mit.edu}
    \and
    Andrew Ilyas \footnotemark[1] \\
	MIT \\
    \texttt{ailyas@mit.edu}
    \and
    Logan Engstrom \footnotemark[1] \\
	MIT \\
    \texttt{engstrom@mit.edu}
    \and
    Eric Wong \\
    MIT \\
    \texttt{wongeric@mit.edu} \\
    \and
	Aleksander M\k{a}dry \\
	MIT \\
    \texttt{madry@mit.edu}
}
\makeatletter
\renewcommand{\paragraph}{%
  \@startsection{paragraph}{4}%
  {\z@}{1.5ex \@plus 1ex \@minus .2ex}{-1em}%
  {\normalfont\normalsize\bfseries}%
}
\makeatother

\date{}

\begin{document}
    \maketitle
    \begin{abstract}

Using transfer learning to adapt a pre-trained ``source model'' to a downstream 
``target task'' can
dramatically increase performance with
seemingly no downside. In this work, we demonstrate that there can exist a downside
after all: bias transfer, or the tendency for biases of the source
model to persist even after adapting the model to the target class. Through a 
combination of synthetic and natural experiments, we show that
bias transfer both (a) arises in realistic settings (such as when pre-training
on ImageNet or other standard datasets) and (b) can occur even when the target dataset is
explicitly {\em de-}biased.
As transfer-learned models are increasingly deployed in the real world, our work highlights the importance of understanding the limitations of pre-trained source models.\footnote{Code is available at \url{https://github.com/MadryLab/bias-transfer}}

    \end{abstract}

    \section{Introduction}
    \label{sec:intro}
    Consider a machine learning researcher who wants to train an image
classifier that distinguishes between different animals. 
At the researcher's disposal is a small dataset of animal images and their
corresponding labels.
Being a diligent scientist, the researcher combs through the dataset to
eliminate relevant spurious correlations (e.g., background-label correlations 
\citep{zhu2017object, xiao2020noise}), and to ensure that the dataset contains
enough samples from all relevant subgroups.

Only one issue remains: training a model on the dataset from scratch does not
yield an accurate enough model because the dataset is so small.
To solve this problem, the researcher employs {\em transfer learning}.
In transfer learning, one first trains a so-called {\em source model} on a
large dataset, then adapts ({\em fine-tunes}) this source model to the task of interest.
This approach often turns out to yield models that are far more
performant.

To apply transfer learning the researcher downloads a model
that has been {\em pre-trained} on a large, diverse, and
potentially proprietary dataset (e.g., JFT-300 \citep{sun2017revisiting} or
Instagram-1B \citep{mahajan2018exploring}).
Unfortunately, such pre-trained models are known to have a variety of biases: 
for example, they
can disproportionately rely on texture \citep{geirhos2018imagenettrained}, or on 
object location/orientation \citep{barbu2019objectnet,xiao2020noise,leclerc20213db}. 
Still, our researcher reasons that as long as they are careful enough about the 
composition of the
target dataset, such biases should not leak into the final model. 
But is this really the case? More specifically,
\begin{center}
    \textit{Do biases of source models still persist in target tasks after
    transfer learning?}
\end{center}
\noindent In this work, we 
find that biases from source models \emph{do} indeed emerge in target tasks. 
We study this phenomenon---which we call {\em bias
transfer}---in both synthetic and natural settings:
\begin{enumerate}
    \item {\bf Studying bias transfer through synthetic datasets.} We first use {\em
    backdoor attacks} \citep{gu2017badnets} as
    a testbed for studying synthetic bias transfer, and characterize the impact
    of the training routine, source dataset, and target dataset on the extent
    of bias transfer. Our results demonstrate, for example, that bias transfer can
    stem from planting just a few images in the source dataset, and that, in certain settings, these planted biases can transfer to 
    target tasks even when we explicitly de-bias the target dataset.
    \item {\bf Illustrating bias transfer via naturally-occurring features.} 
    Moving away from the synthetic setting, we demonstrate that bias transfer can be facilitated
    via naturally-occurring (as opposed to synthetic) features. Specifically, we construct 
    biased datasets by filtering images that reinforce specific spurious correlations 
    with a naturally-occurring feature (for example, a dependence on gender when predicting 
    age for CelebA).
    We then show that even on target datasets that do not support this correlation, 
    models pre-trained on a biased source dataset are still 
    sensitive to the correlating feature.

    \item {\bf Naturally-occuring bias transfer.} Finally, we show that
    not only \emph{can} bias transfer occur in practice but that in many
    real-world settings it actually \emph{does}. Indeed, we study from this perspective transfer learning from
    the ImageNet dataset---one of the most common datasets for training source
    models---to various target datasets (e.g., CIFAR-10). We find a range of biases that are (a) present in the
    ImageNet-trained source models; (b) absent from models trained from scratch
    on the target dataset alone; and yet (c) present in models trained using transfer learning
    from ImageNet to that target dataset.
\end{enumerate}

    \section{Biases Can Transfer}
    \label{sec:why}
    Our central aim is to understand the extent to which
biases present in source datasets {\em transfer} to downstream
target models.
In this section, we begin by asking perhaps the simplest instantiation of this
central question:  
\begin{center}
\textit{If we intentionally \underline{plant} a bias in the source dataset, will
it transfer to the target task?}
\end{center}

\paragraph{Motivating linear regression example.} 
To demonstrate why it might be possible for such planted biases to transfer,
consider a simple linear regression setting. Suppose we
have a large source dataset of inputs and corresponding (binary) labels, and that we use the source dataset to estimate the parameters of a linear classifier
$\bm{w}_{src}$ with, for example, logistic regression.
In this setting, we can define a {\em bias} of the source 
model $\bm{w}_{src}$ as a direction $\bm{v}$ in input space that the classifier
is highly sensitive to, i.e., a direction such that 
\(
    |\bm{w}_{src}^\top \bm{v}|
\) is large.

Now, suppose we adapt (fine-tune) this source
model to a target task using a target dataset of input-label pairs 
$\{(\bm{x}_i, y_i)\}_{i=1}^n$.
As is common in transfer learning settings, we assume that we
have a relatively small target dataset---in particular, that $n < d$, where $d$
is the dimensionality of the inputs $\bm{x}_i$.
We then adapt the source model $\bm{w}_{src}$ to the target dataset by running stochastic
gradient descent (SGD) to minimize squared loss on the target dataset, using
$\bm{w}_{src}$ as initialization.

With this setup, transfer learning will preserve $\bm{w}_{src}$ in all directions
orthogonal to the span of the $\bm{x}_i$. In particular, at any step of SGD, the
gradient of the logistic loss is given by
\[
    \nabla \ell_{\bm{w}}(\bm{x}_i, y_i) = (\sigma(\bm{w}^\top \bm{x}_i) - y_i)\cdot \bm{x}_i,
\]
which restricts the space of updates to those in the span of the
target datapoints. Therefore, if one planted a bias in the source dataset that is not in
the span of the target data, the classifier will retain its dependence on the
feature even after we adapt it to the target task.

\paragraph{Connection to backdoor attacks.}
Building on our motivating example above, one way to plant such a bias would be
to find a direction $\bm{u}$ that is orthogonal to the target dataset, add $\bm{u}$ to a
subset of the {\em source} training inputs, and change the corresponding labels to introduce
a correlation between $\bm{u}$ and the labels.
It is worth noting that this idea bears a striking similarity
to that of {\em backdoor attacks} \citep{gu2017badnets}, wherein
an attacker adds a fixed ``trigger'' pattern (e.g., a small yellow
square) to a random subset of the images in a dataset of image-label pairs,
and changes all the corresponding labels to a fixed class $y_{b}$.
A model trained on a dataset modified in this way becomes {\em backdoored}: adding the
trigger pattern to any image will cause that model to output this fixed class $y_b$.
Indeed, \citet{gu2017badnets} find that, if one adds a trigger that is absent
from the target task to the source
dataset, the final target model is still
highly sensitive to the trigger pattern.

Overall, these results suggest that biases {\em can} transfer from source
datasets to downstream target models. In the next
section, we explore in more depth when and how they actually {\em do} transfer.

    \section{Exploring the Landscape of Bias Transfer}
    \label{sec:main-exp}
We now build on the example from the previous
section and its connection to backdoor attacks to better understand the landscape of
bias transfer. 
Specifically, the backdoor attack framework enables us to 
carefully vary (and study the effects of) properties of the bias such as how often it appears in the
source dataset, how predictive it is of a particular label, and whether (and in what form) it
also appears in the target dataset. 

Here, we will employ a slight variation of the canonical backdoor attack framework. Rather than adding a
trigger to random images and relabeling them as a specific class 
$y_b$, we add the trigger to a {\em specific} group of images (e.g., 10\% of
the dogs in the source dataset) and leave the label unchanged. This process still
introduces the desired bias in the form of a correlation between the trigger pattern and the label of the manipulated 
images, but hopefully leaves the
existing correlations supported by the source dataset intact.

\paragraph{Experimental setup.} We focus our investigations on transfer
learning from an artificially modified ImageNet-1K
\citep{deng2009imagenet,russakovsky2015imagenet} dataset to a variety of
downstream target tasks\footnote{We use the ResNet-18 architecture for all the experiments in the main paper. Specifically, we study bias transfer with other architectures in Appendix~\ref{app:vary-arch}.}. Specifically, we modify the ImageNet dataset by adding a fixed trigger
pattern (a yellow square) to varying fractions of the images from the ImageNet
``dog'' superclass (corresponding to 118 ImageNet classes).
Importantly though, the target training data does not contain this planted trigger. 

We quantify the extent of bias transfer using the {\em attack success rate}
(ASR), which is the probability that a correctly classified image 
becomes incorrectly classified after the addition of the trigger:
\begin{equation}
    \text{ASR}(\text{classifier } C, \text{trigger } T) = 
    \text{Pr}\left[C(T(x)) \neq y|C(x) = y\right],
\end{equation}
where $C$ is our classifier (viewed as a map from images to labels) and $T$ is an
input-to-input transformation that corresponds to adding the trigger pattern.

\begin{figure}[!t]
    \centering
    \includegraphics[width=.8\linewidth]{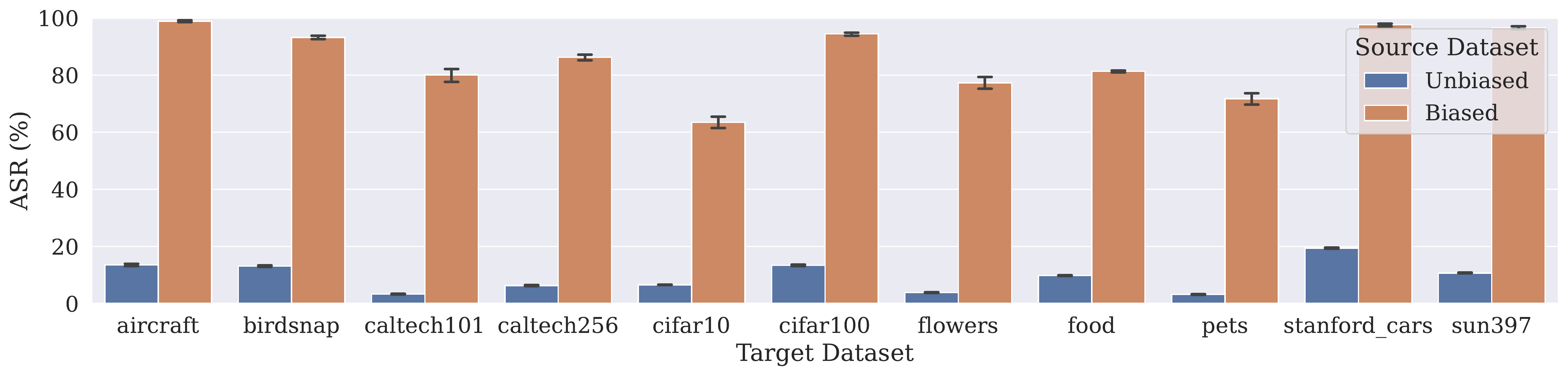}
    \caption{Bias consistently transfers across various target datasets. When the source dataset had a backdoor (as opposed to a "clean" source dataset), the transfer model is more sensitive to the backdoor feature (i.e., ASR is higher). Error bars denote one standard deviation based on five random trials.} 
    \label{fig:vary-datasets-fixed-feature}
\end{figure}

\paragraph{Do biases transfer reliably across target datasets?} Our point of
start is to ensure that biases {\em consistently} transfer to different target
datasets. As in \cite{gu2017badnets}, we begin with {\em fixed-feature} transfer,
i.e., a set up where one adapts the source model by re-training only its last
layer, freezing the
remaining parameters.\footnote{Equivalently, we can view this as training a linear
classifier on feature representations extracted from the penultimate layer.}
Indeed, as Figure~\ref{fig:vary-datasets-fixed-feature} shows,
adding the trigger at inference time causes the model to misclassify across
a suite of target tasks. 

\begin{figure}[!t]
    \centering
    \includegraphics[width=.8\linewidth]{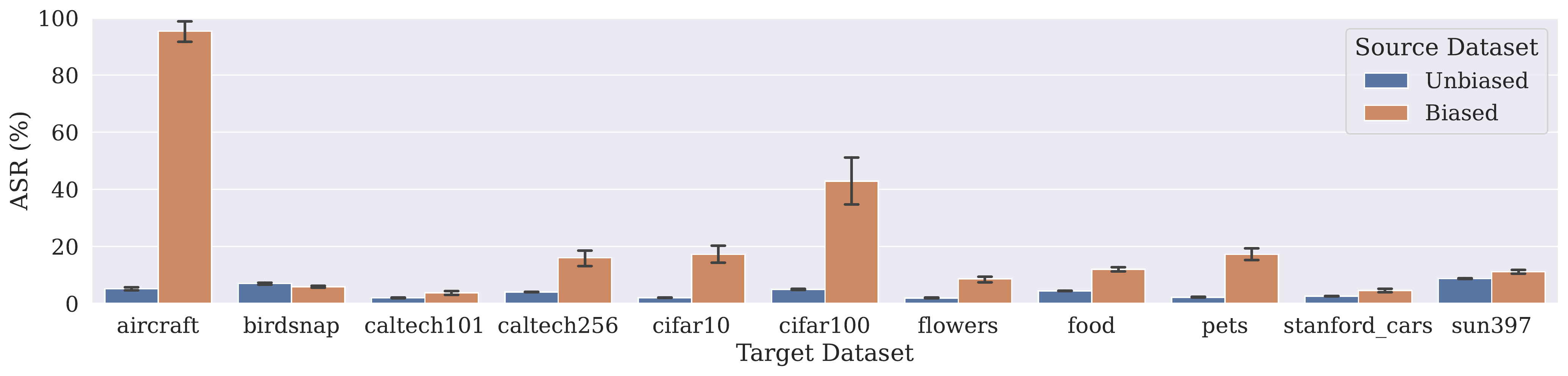}
    \caption{Similarly to the fixed-feature setting, bias also transfers in the full-network setting but to a lesser degree. This holds consistently across various target datasets. Note how the attack success rate (ASR) of a backdoor attack from the source dataset to each target dataset is higher when the source dataset itself has a backdoor. Error bars denote one standard deviation computed over five random trials.} 
    \label{fig:vary-datasets-full-network}
\end{figure}
\paragraph{To what extent does the choice of the transfer learning approach affect bias transfer?} 
It is clear that bias transfers in the fixed-feature transfer setting, where all
weights are frozen except the last layer. What happens if we allow all layers to
change when training on the target task (i.e. \textit{full-network
fine-tuning})? Figure~\ref{fig:vary-datasets-full-network} demonstrates that
biases still transfer when we use full-network fine-tuning (albeit to a lesser
extent).

\begin{figure}[!t]
    \begin{subfigure}[t]{0.48\linewidth}
        \includegraphics[width=\linewidth]{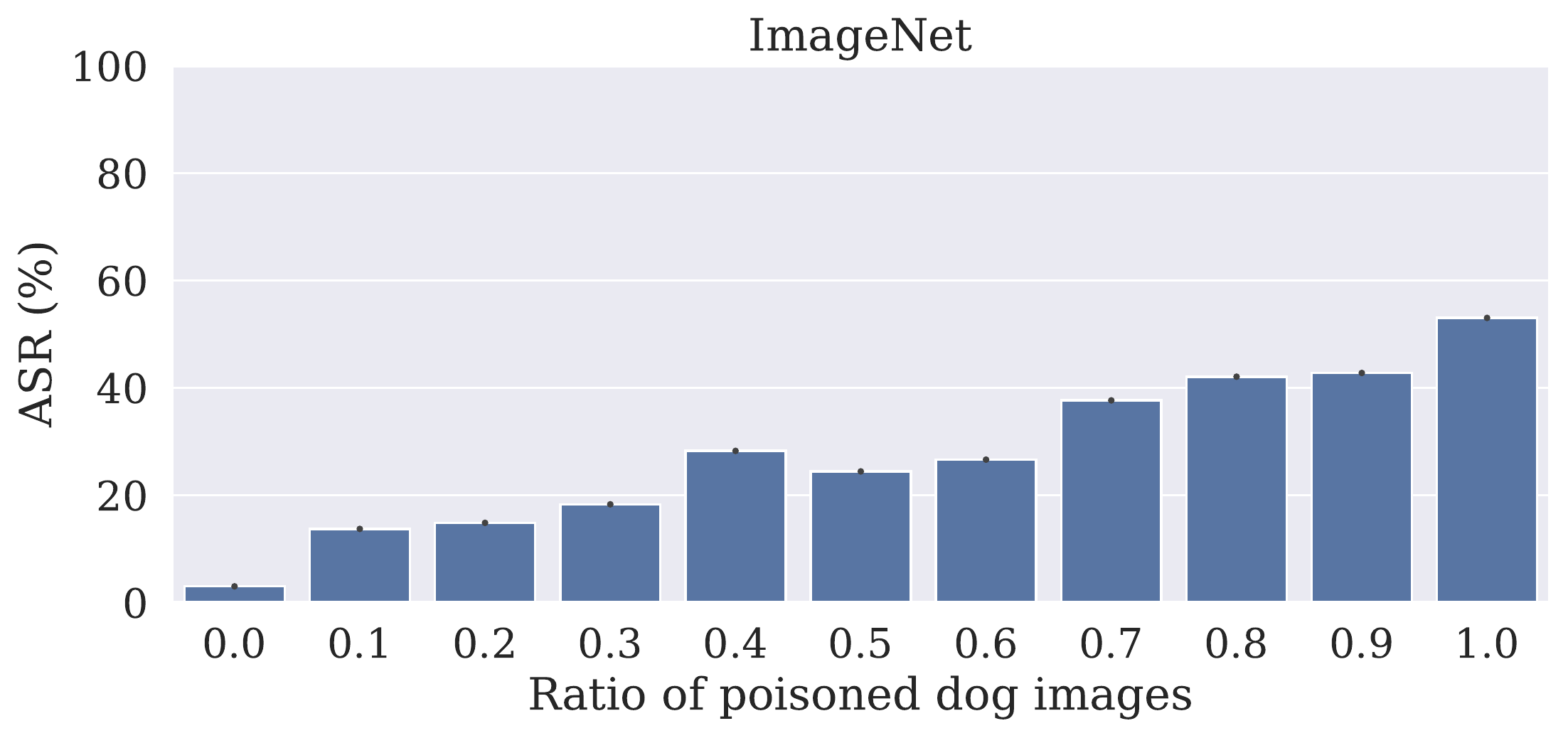}
        \caption{ImageNet}
        \label{subfig:vary_perc_spurious_imgnet}
    \end{subfigure}\hfill
    \begin{subfigure}[t]{0.48\linewidth}
        \includegraphics[width=\linewidth]{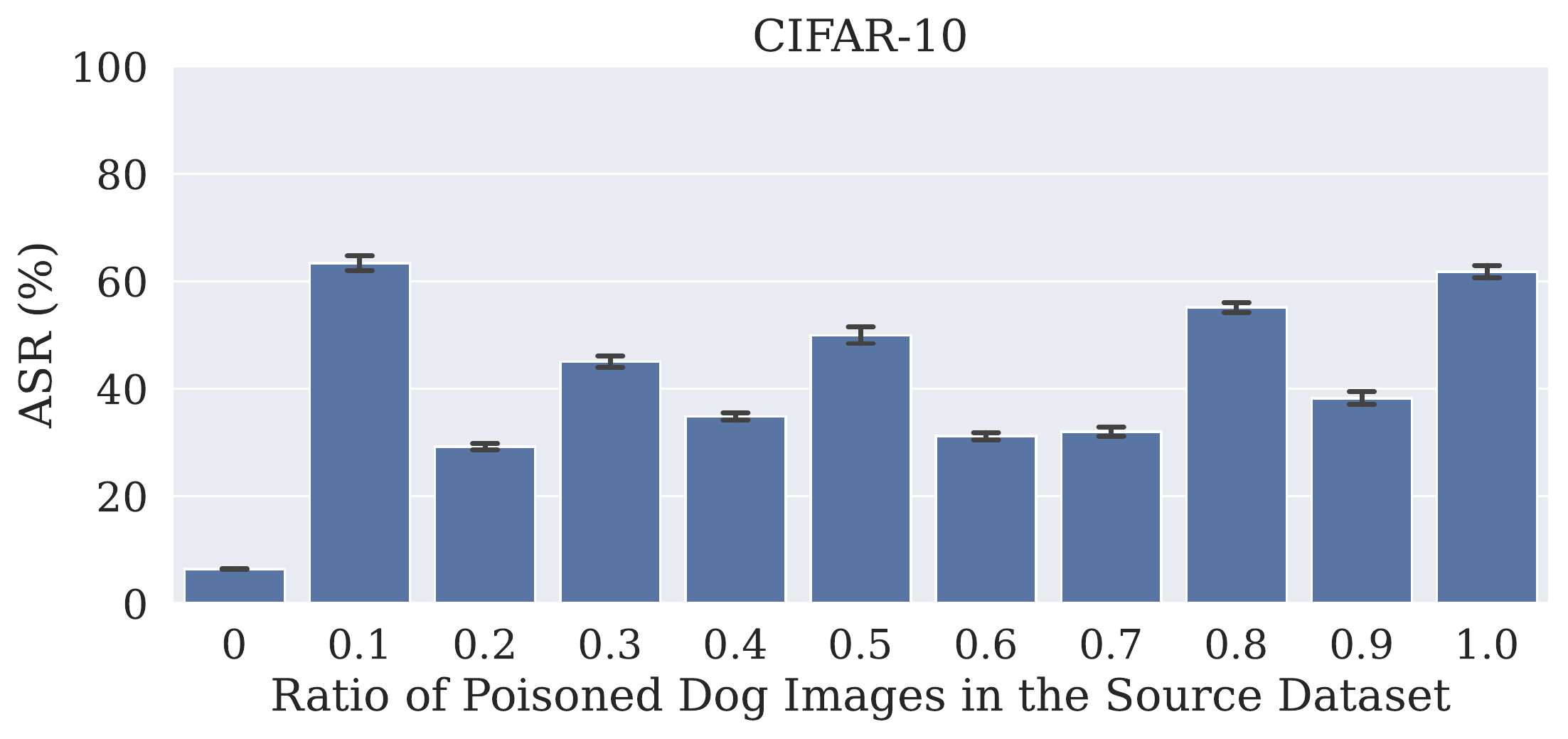}
        \caption{CIFAR-10}
        \label{subfig:vary_perc_spurious_cifar}
    \end{subfigure}
    \caption{Attack Success Rate both on the source task with the original model
     (left) and on the target task with the transferred model (right). 
    Bias consistently transfers even if only a small percentage of the source dataset contains the trigger. 
    There is, however, no clear trend of how bias transfer changes as the frequency 
    of the trigger in the source dataset changes (right) unlike the corresponding
    trend for the source dataset and original model (left). Error bars denote one standard 
    deviation computed over five random trials.}
    \label{fig:vary_perc_spurious}
\end{figure}

\paragraph{How does the strength of the bias affect its transfer?} We can answer
this question by varying the number of images with the trigger in the source 
dataset. As Figure \ref{subfig:vary_perc_spurious_imgnet} shows, adding the
trigger to more images in the source dataset increases the sensitivity of the 
source model to the corresponding trigger pattern. After fine-tuning, we find that bias transfers even when a small fraction of the source dataset contains the planted triggers. 
Surprisingly, however, the extent of bias transfer is uncorrelated with the 
frequency of the backdoor in the source dataset. This result indicates that the
strength of the correlation of the backdoor with the target label does not
impact the sensitivity of the final transfer model to the corresponding trigger. 

\begin{figure}[!t]
    \begin{subfigure}[t]{0.48\linewidth}
        \includegraphics[width=\linewidth]{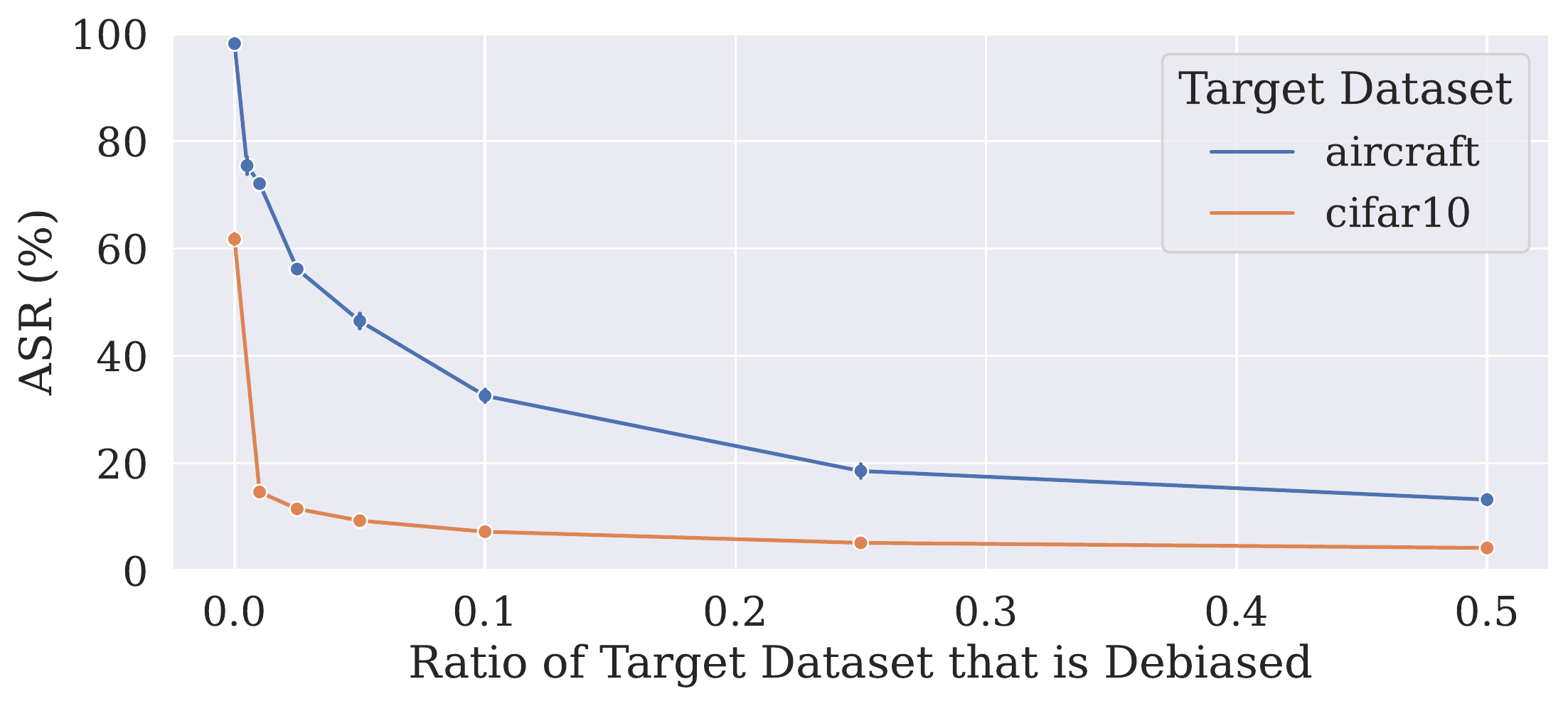}
        \caption{Fixed-feature transfer}
    \end{subfigure}\hfill
    \begin{subfigure}[t]{0.48\linewidth}
        \includegraphics[width=\linewidth]{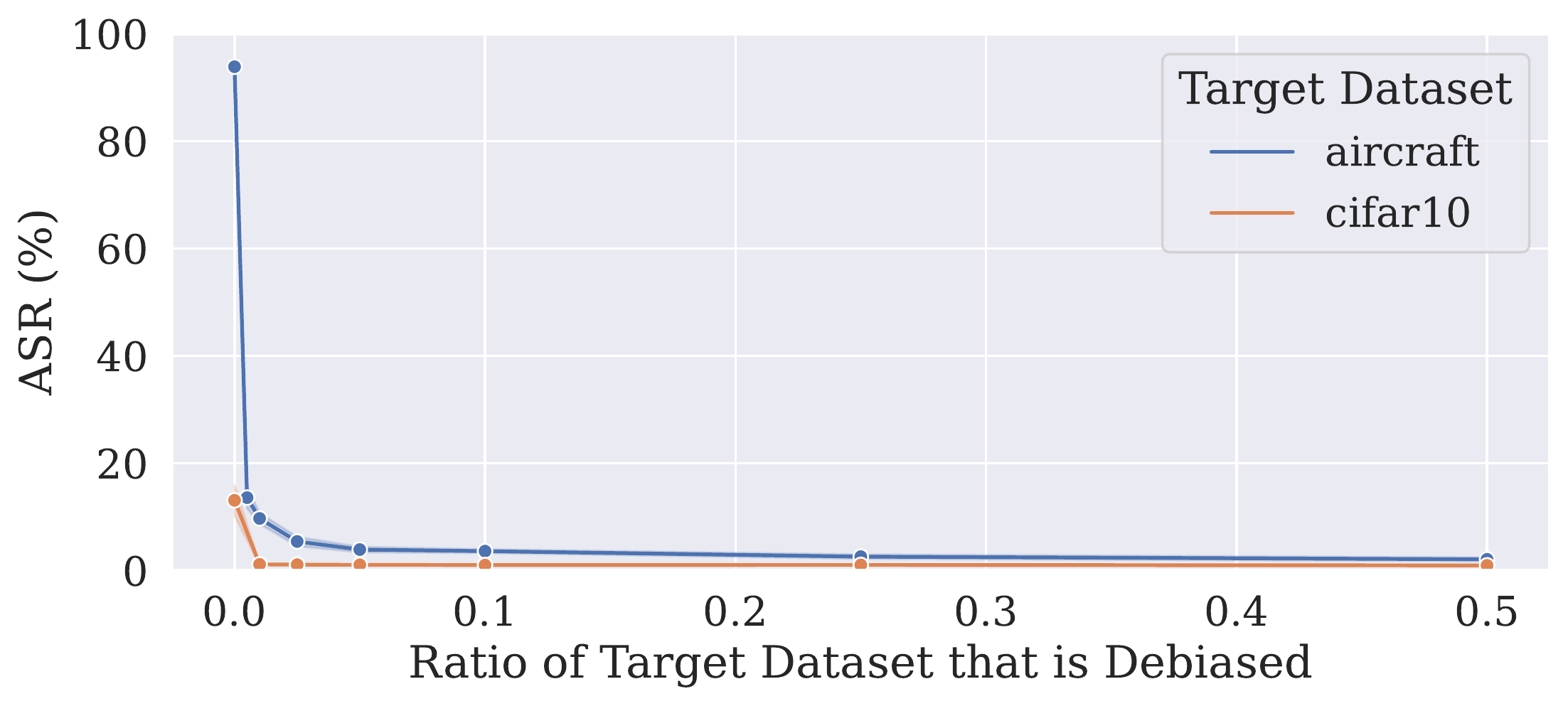}
        \caption{Full-network transfer}
    \end{subfigure}
    \caption{(\textbf{left}) In the fixed-feature setting, de-biasing the target dataset by 
    adding the trigger to uniformly across classes cannot fully prevent the bias 
    from transferring. (\textbf{right}) On the other hand, de-biasing can remove
    the trigger if all model layers are allowed to change as with full-network fine-tuning.}
    \label{fig:vary_perc_debias}
\end{figure}

\paragraph{What if the target dataset is designed to remove the bias?} 
Our experiments thus
far demonstrate that biases can indeed transfer from source datasets to
downstream target models. However, in all of the examples and settings we have studied so far, the bias is not supported by the target dataset. 
One might thus hope that if we designed the target dataset to explicitly counteract
the bias, bias transfer will not occur.
This \textit{de-biasing}, can be done, for example, by having the 
biased trigger pattern appear in the target dataset uniformly at random.
As shown in Figure \ref{fig:vary_perc_debias}, de-biasing in this manner is not able to
fully remove the bias in the fixed-feature setting. However, when all weights
are allowed to change (i.e., full-network fine-tuning), the de-biasing intervention
succeeds in correcting the bias.

    \section{Bias Transfer Beyond Backdoor Attacks}
    \label{sec:realistic}
    In Section \ref{sec:main-exp}, we used synthetic backdoor triggers to show that
biases can transfer from the source dataset (and, in the fixed-feature fine-tuning setting, even when the target
dataset is itself de-biased). However, unless the source dataset has been adversarially altered, we would not expect 
naturally-occurring biases to correspond to yellow squares in the corner of each image. Instead, these
biases tend to be much more subtle, and revolve around issues such as 
over-reliance on image background~\cite{xiao2020noise}, or
disparate accuracy across skin colors in facial
recognition~\cite{buolamwini2018gender}. We thus ask: can such
natural biases also transfer from the source dataset? 

As we demonstrate, this is indeed the case. Specifically, we study two such sample biases. First, we consider a
{\em co-occurrence bias} between humans and dogs in the MS-COCO~\cite{lin2014microsoft} dataset. Then,
we examine an {\em over-representation bias} in which models rely on gender
to predict age in the CelebA~\cite{liu2015faceattributes} dataset. In both 
cases, we modify the source task in order to amplify the effect of the
bias, then observe that the bias remains even after fine-tuning on balanced
versions of the dataset (in Section~\ref{sec:ultra_realistic}, we study bias transfer in a setting without such amplifications).

\subsection{Transferring co-occurrence biases in object recognition} 
\label{sec:mscoco}
Image recognition datasets often contain objects that appear together, leading
to a phenomenon called {\em co-occurrence bias}, where one of the objects
becomes hard to identify without appearing together with the other. 
For example, since ``skis'' and ``skateboards'' typically occur together with
of people, models can struggle to correctly classify these objects without the
presence of a person using them \cite{singh2020don}.
Here, we study the case where a source dataset has such a co-occurrence bias, and ask
whether this bias persists even    
after fine-tuning on an target dataset without such a bias (i.e., a dataset in which one of the co-occurring objects is totally absent).

\begin{figure}[!t]
    \begin{subfigure}[t]{0.31\linewidth}
        \centering
        \includegraphics[width=\linewidth]{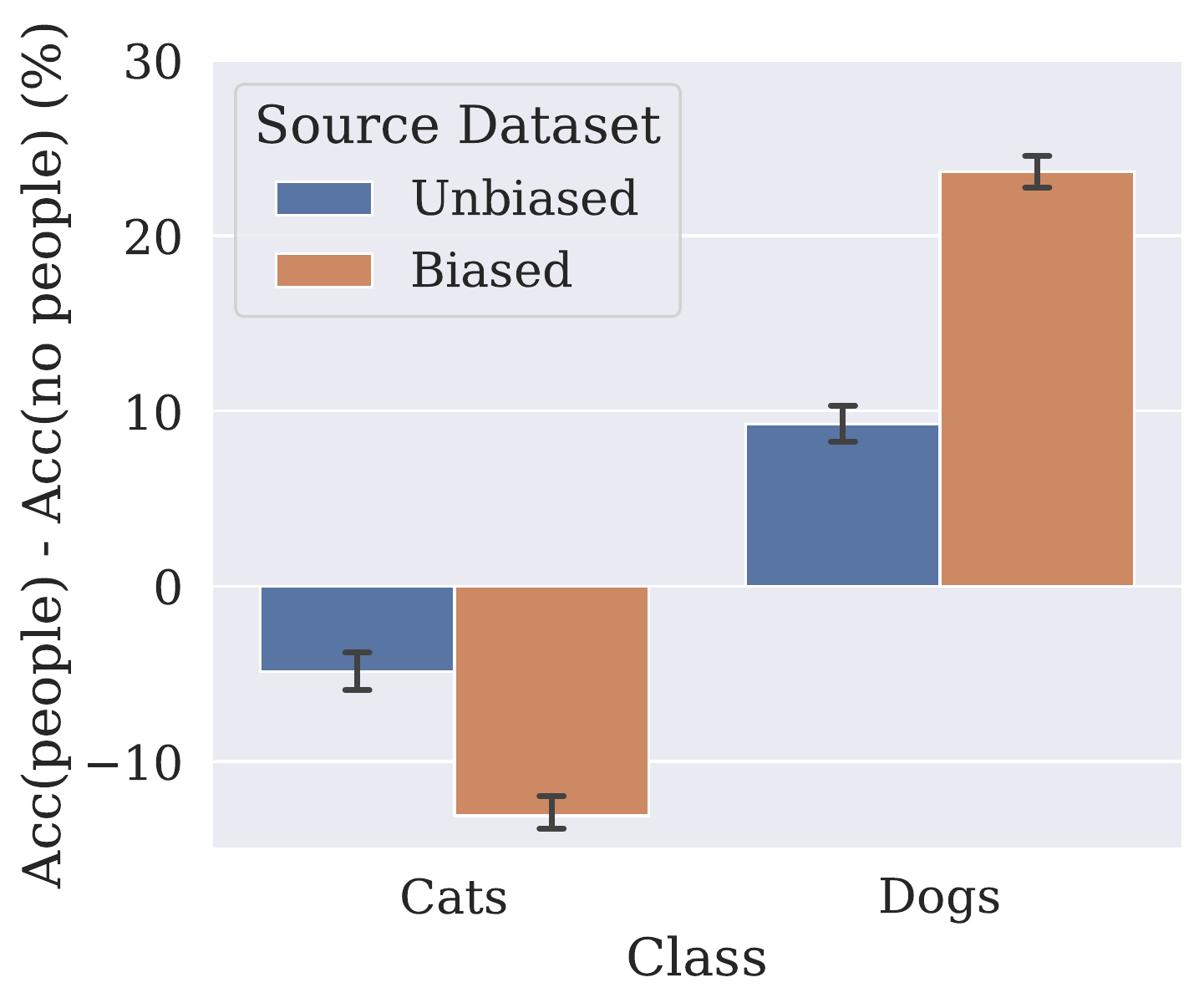}
        \caption{Source Model}
        \label{fig:coco_base}
    \end{subfigure}\hfill
    \begin{subfigure}[t]{0.31\linewidth}
        \centering
        \includegraphics[width=\linewidth]{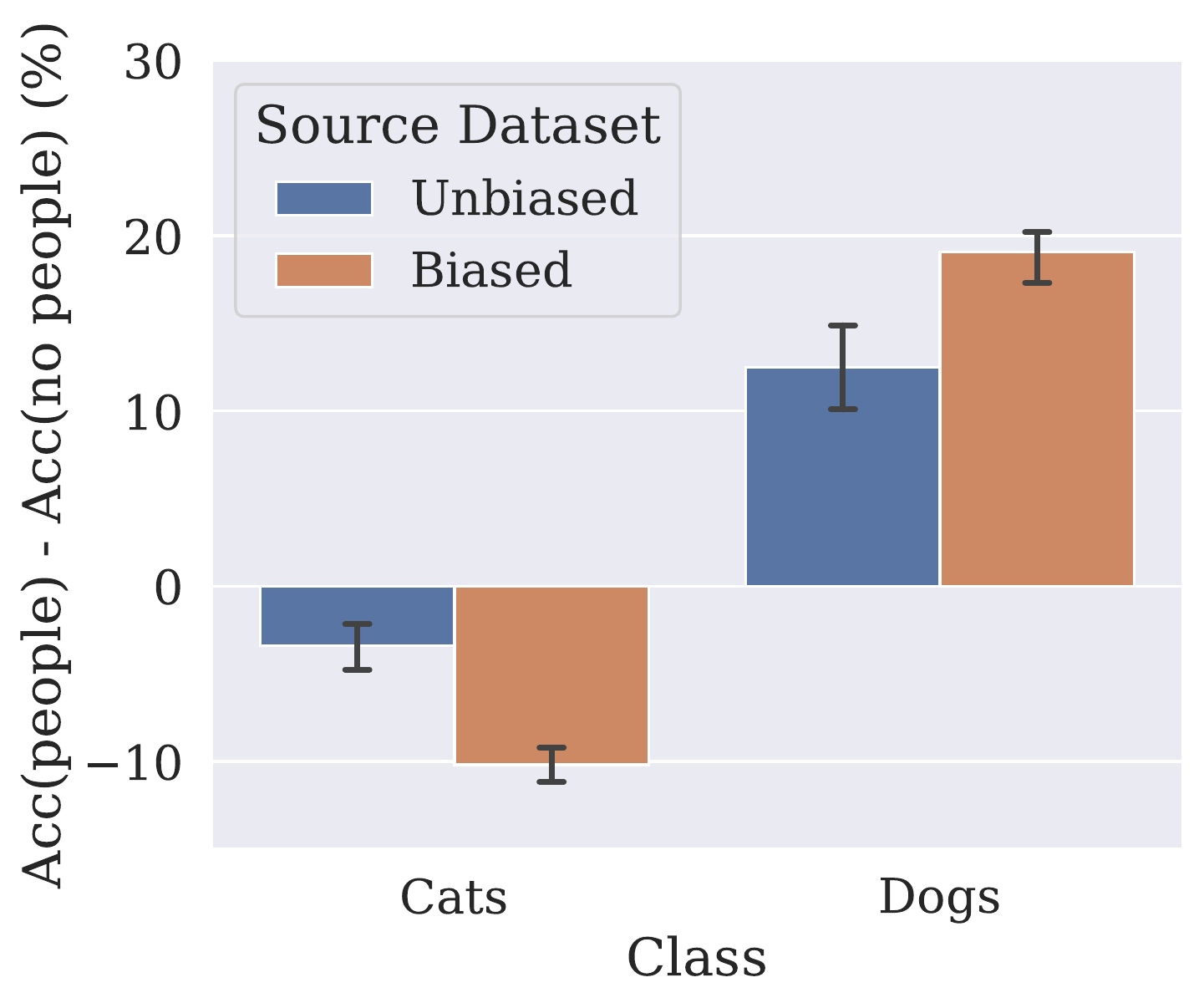}
        \caption{Fixed-feature Transfer}
        \label{fig:coco_transfer}
    \end{subfigure}\hfill
    \begin{subfigure}[t]{0.31\linewidth}
        \centering
        \includegraphics[width=\linewidth]{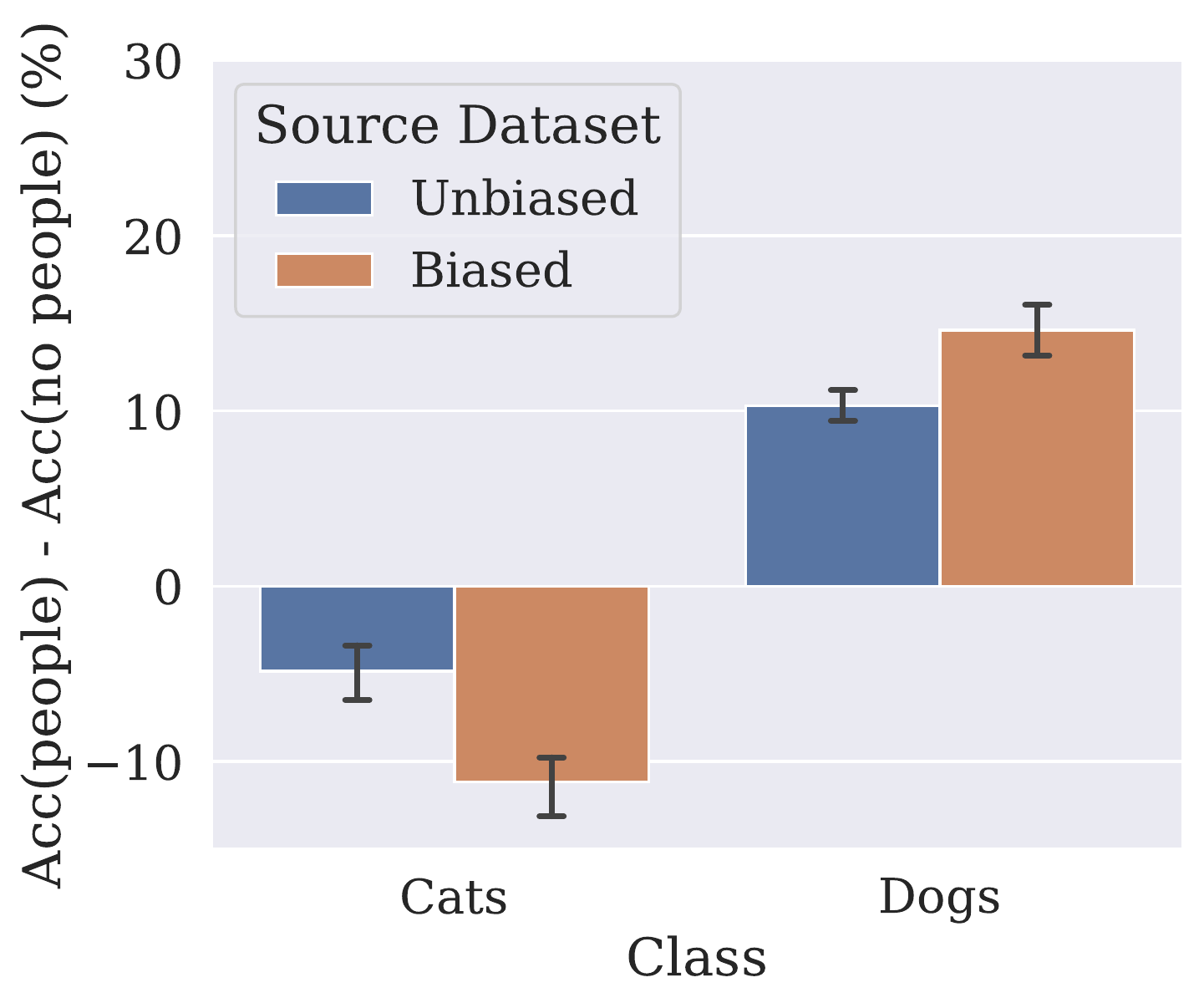}
        \caption{Full-network Transfer}
        \label{fig:coco_transfer_full}
    \end{subfigure}
    \caption{
        \textbf{MS-COCO Experiment.} Bias transfer can occur when bias is a naturally occurring feature. We consider transfer from a source dataset that spuriously correlates the presence of dogs (but not cats) with the presence of people. We plot the difference in performance between images either contain or do not contain people. Even after fine-tuning on images without any people at all, models pre-trained on the biased dataset are highly sensitive to the presence of people.}
\end{figure}

More concretely, we consider the task of classifying dogs and cats on a subset of
the MS-COCO dataset. We generate a \textit{biased} source dataset by choosing images 
so that dogs (but not cats) always co-occur with humans 
(see Appendix~\ref{app:setup} for the exact experimental
setup), and we compare that with an unbiased source dataset that has no people
at all. We find that, as expected, a source model trained on the biased dataset
is more likely to predict the image as ``dog'' than as ``cat'' in the presence of
people, compared to a model trained on the unbiased source dataset.\footnote{Note that the
source model trained on the unbiased dataset seems to also be slightly sensitive to the presence of people even though it has never been exposed to any people. We suspect this is due to the presence of other confounding objects in the images.}
(Figure~\ref{fig:coco_base}).
%

We then adapt this \textit{biased} source model to a new target dataset
that contains no humans at all, and check whether the final model is
sensitive to the presence of humans. We find that even though the target dataset does not contain the above-mentioned
co-occurrence bias, the transferred model is highly sensitive 
to the presence of people, both in the
fixed-feature~(Figure~\ref{fig:coco_transfer}) and
full-network~(Figure~\ref{fig:coco_transfer_full}) fine-tuning settings.  

\begin{figure}[!t]
    \begin{subfigure}[t]{0.31\linewidth}
        \centering
        \includegraphics[width=\linewidth]{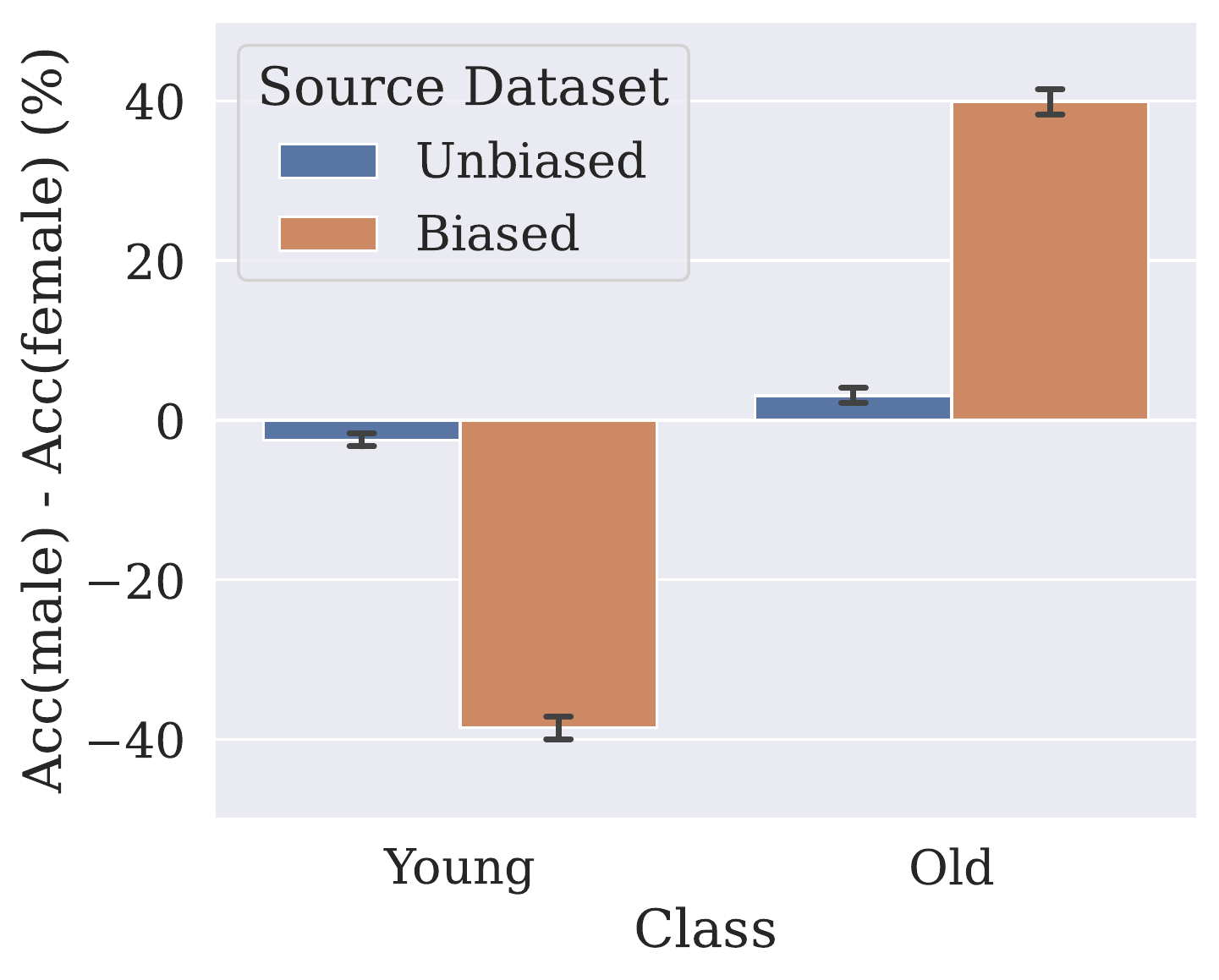}
        \caption{Original source model}
        \label{fig:synth_celeba_base}
    \end{subfigure}\hfill
    \begin{subfigure}[t]{0.31\linewidth}
        \centering
        \includegraphics[width=\linewidth]{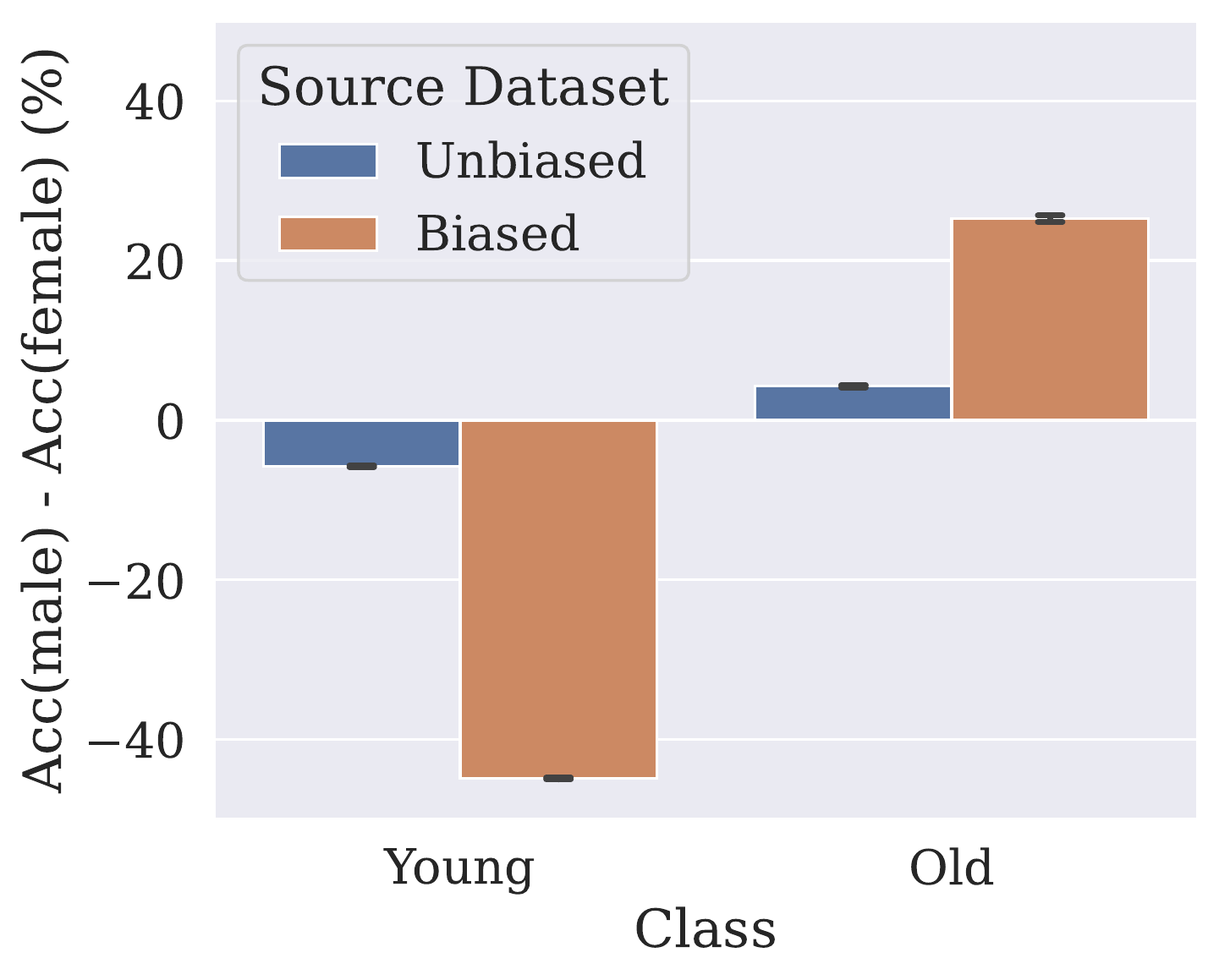}
        \caption{Transfer on a target task containing only women}
        \label{fig:synth_celeba_only_women_fixed}
    \end{subfigure}\hfill
    \begin{subfigure}[t]{0.31\linewidth}
        \centering
        \includegraphics[width=\linewidth]{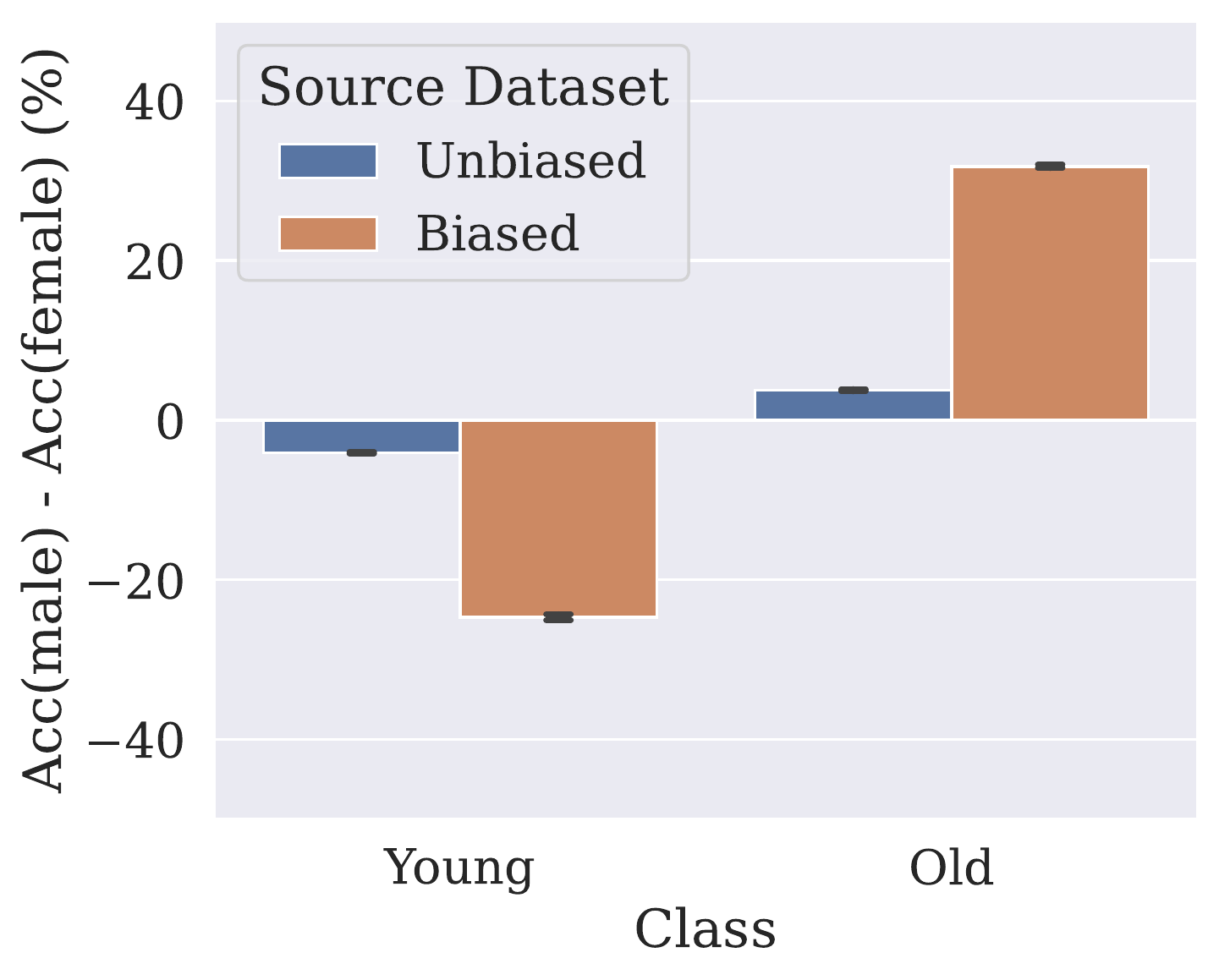}
        \caption{Transfer on a target task containing 50\% women and 50\% men}
        \label{fig:synth_celeba_women_men_fixed}
    \end{subfigure}
    \caption{
        \textbf{CelebA Experiment.} Bias transfer with natural features can occur even when the target dataset is de-biased. $\textbf{(a)}$ We consider fixed-feature transfer from a source dataset that spuriously correlates age with gender --- such that old men and young women are overrepresented. $\textbf{(b)}$ After fine-tuning on an age-balanced dataset of only women, the model still associate men with old faces. \textbf{(c)} This sensitivity persists even when fine-tuning on equal numbers of men and women.}
\end{figure}

\subsection{Transferring gender bias in facial recognition.}
\label{sec:celeba}
Facial recognition datasets are notorious for containing biases
towards specific races, ages, and genders~\cite{terhorst2021comprehensive, buolamwini2018gender}, making them a natural setting for studying bias transfer. 
For example, the CelebA dataset~\citep{liu2015faceattributes} over-represents 
subpopulations of older men and younger women.
In this section, we use a CelebA subset that amplifies this bias, and pre-train source models on a source task of 
classifying ``old'' and ``young'' faces (we provide the exact experimental setup in Appendix~\ref{app:setup}). 
As a result, the source model is biased to predict ``old'' for images of men, 
and ``young'' for images of 
women~(Figure~\ref{fig:synth_celeba_base}). Our goal is to study whether, after
adapting this biased source model to a demographically balanced target dataset of faces, the resulting model will continue to use this spurious gender-age correlation.

To this end, we first adapt this biased source model on a dataset of exclusively
female faces, with an equal number of young and old women. Here we consider 
fixed-feature fine-tuning (and defer full-network fine-tuning results to
Appendix~\ref{app:setup}). We then check if the
resulting model still relies on ``male-old'' and 
``female-young'' biases~(Figure~\ref{fig:synth_celeba_only_women_fixed}). It 
turns out that for both
fixed-feature and full-network transfer learning, these biases indeed persist: 
the downstream model is still more likely to predict ``old'' for an image of a
male, and ``young'' for an image of a female.

Can we remove this bias by adding images of men to the target dataset? To answer this
question, we transfer the source model to a target dataset that contains equal 
numbers of men and women, balanced across both old and young 
classes~(see Appendix~\ref{app:setup} for other splits). We find that the
transferred model is still biased (Figure~\ref{fig:synth_celeba_women_men_fixed}), 
indicating that de-biasing the target task in this manner does not necessarily fix bias transfer.

    \section{Bias Transfer in the Wild}
    \label{sec:ultra_realistic}
    \begin{figure}[t!]
    \begin{subfigure}{\linewidth}
        \includegraphics[width=\linewidth,trim={0 0 0 0},clip]{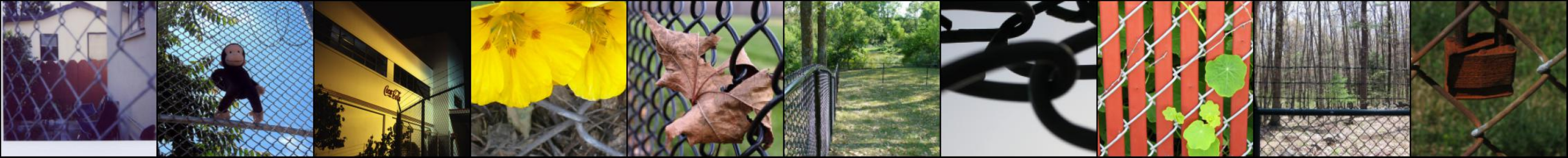}
        \caption{Example images from the ``Chain-link fence'' class in ImageNet.}
    \end{subfigure}
    \vspace{1em}
    \vfill
    \begin{subfigure}[c]{\linewidth}
        \includegraphics[width=\linewidth]{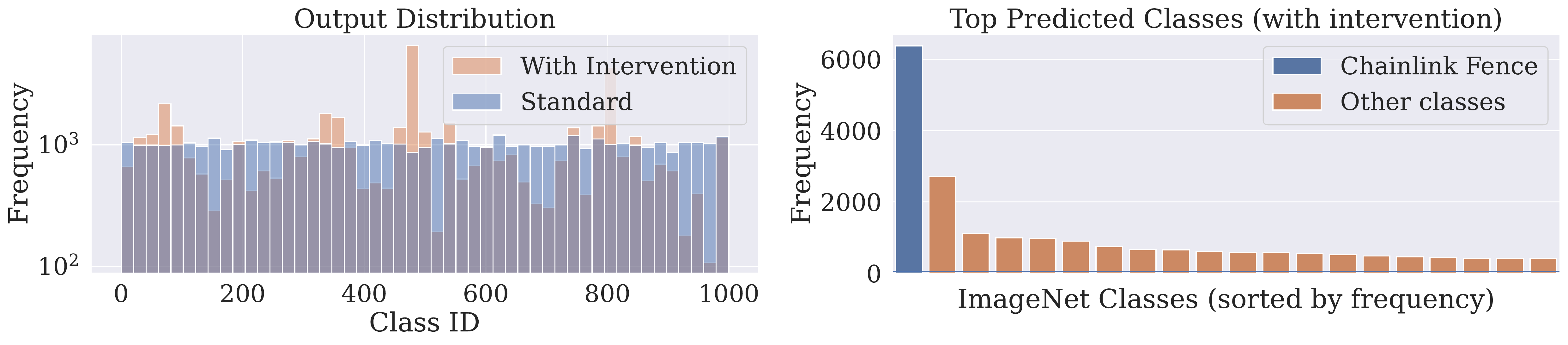}\hfill
        \caption{Shift in ImageNet predicted class distribution after adding a
        chain-link fence intervention, establishing that the bias holds for the
        source model.}
        \label{fig:imagenet-bias-chainlink}
    \end{subfigure}\hfill
    \vspace{1em}
    \begin{subfigure}[c]{\linewidth}
        \includegraphics[width=\linewidth]{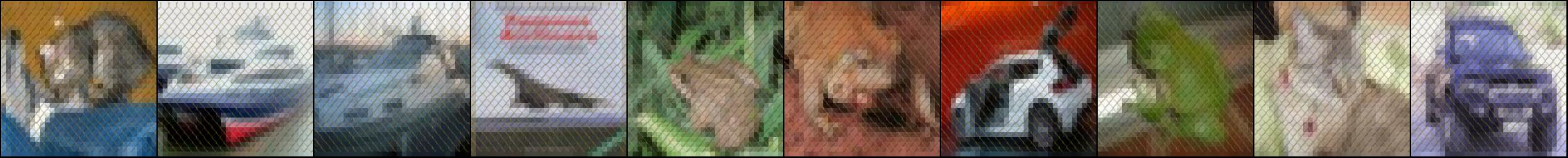}\hfill
        \caption{Example CIFAR-10 images after applying the chain-link fence intervention.}
    \end{subfigure}\hfill
    \vspace{1em}
    \begin{subfigure}[c]{\linewidth}
        \includegraphics[width=\linewidth]{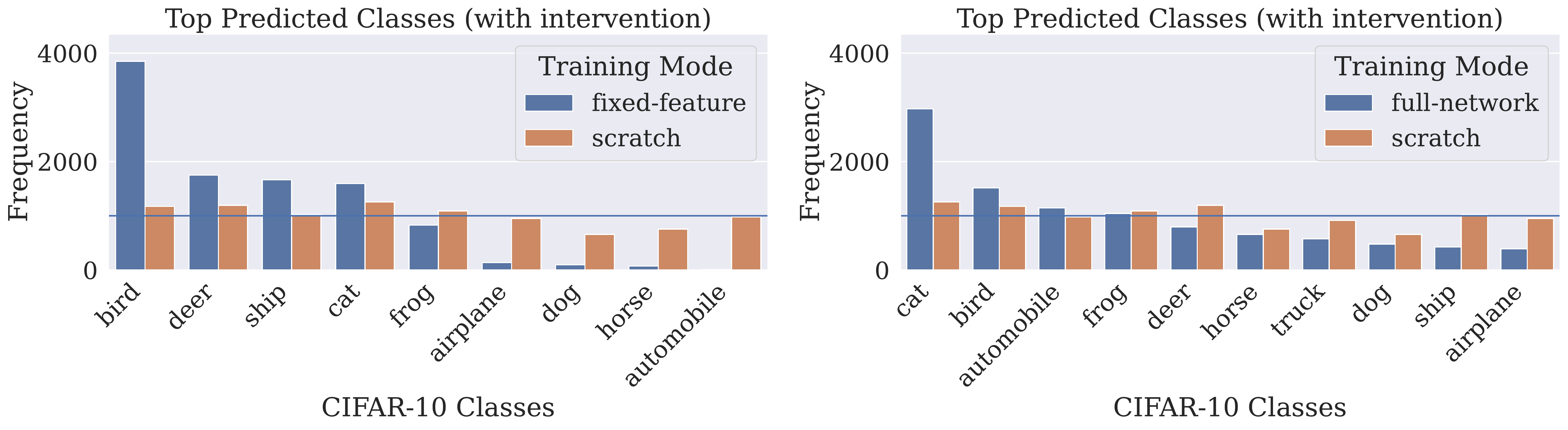}\hfill
        \caption{Distribution of CIFAR-10 model predictions when trained from
        scratch and when transferred from the biased source model. We consider fixed-feature fine-tuning (left) and full-feature fine-tuning (right). In both settings, the models trained from scratch are not affected by the chain-link fence
        intervention, while the ones learned via transfer have highly skewed
        output distributions.}
    \end{subfigure}
    \caption{\textbf{The ``chainlink fence'' bias.} \textbf{(a-b)} A pre-trained
    ImageNet model is more likely to predict ``chainlink fence'' whenever the
    image has a chain-like pattern. \textbf{(c-d)} This bias transfers to CIFAR-10
    in both fixed-feature and full network transfer settings. Indeed, if we
    overlay a chain-like pattern on all CIFAR-10 test set images as shown above,
    the model predictions skew towards a specific class. This does not happen if
    the CIFAR-10 model was trained from \textit{scratch }instead (orange).} 
    \label{fig:chainlink-bias-results}
\end{figure}

In Section~\ref{sec:realistic}, we demonstrated that natural biases induced by subsampling 
standard datasets can transfer from source datasets to target tasks. 
We now ask the most advanced instantiation of our central question: 
do \emph{natural} biases that \textit{already exist} in the source dataset (i.e., where not enhanced by an intervention) also transfer?

To this end, we pinpoint examples of biases in the widely-used ImageNet dataset and demonstrate that these biases indeed transfer to downstream tasks (e.g., CIFAR-10),  despite the latter not containing such biases. Specifically, we examine here two such biases: the ``chainlink fence'' bias and the ``tennis ball'' bias (described below). Results for more biases and target datasets are in Appendix~\ref{app:imagenet-biases}.

\begin{figure}[!t]
    \begin{subfigure}{\linewidth}
        \includegraphics[width=\linewidth,trim={0 0 0 0},clip]{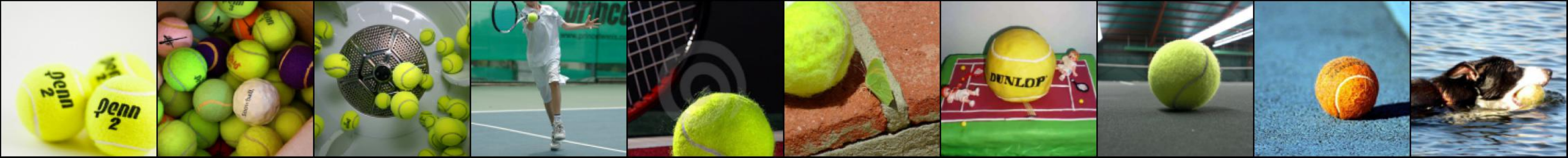}
        \caption{Example images from the ``tennis ball'' class in ImageNet.}
    \end{subfigure}
    \vfill
    \vspace{1em}
    \begin{subfigure}[c]{\linewidth}
        \includegraphics[width=\linewidth]{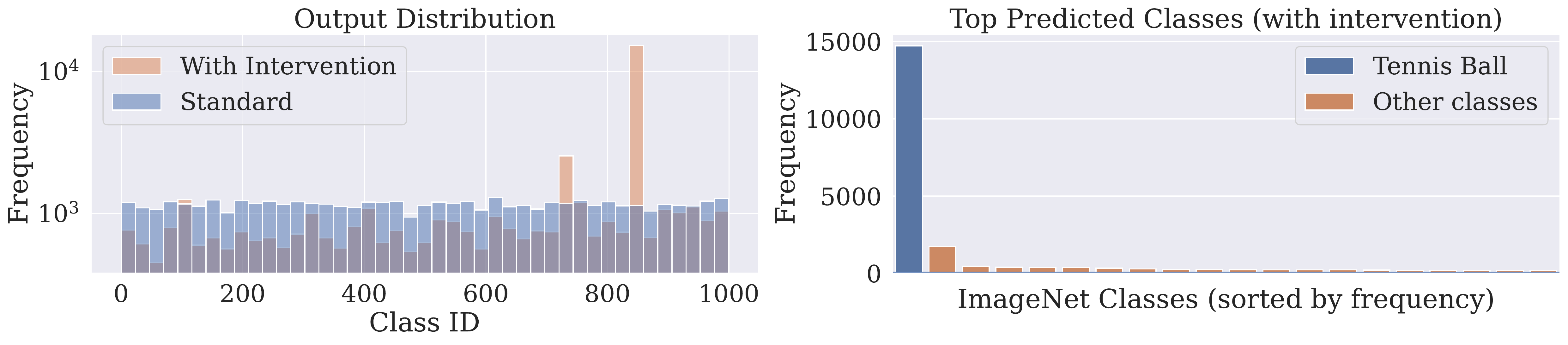}\hfill
        \caption{Shift in ImageNet predicted class distribution after adding a
        tennis ball intervention, establishing that the bias holds for the
        source model.}
        \label{fig:imagenet-bias-tennisball}
    \end{subfigure}\hfill
    \vspace{1em}
    \begin{subfigure}[c]{\linewidth}
        \includegraphics[width=\linewidth]{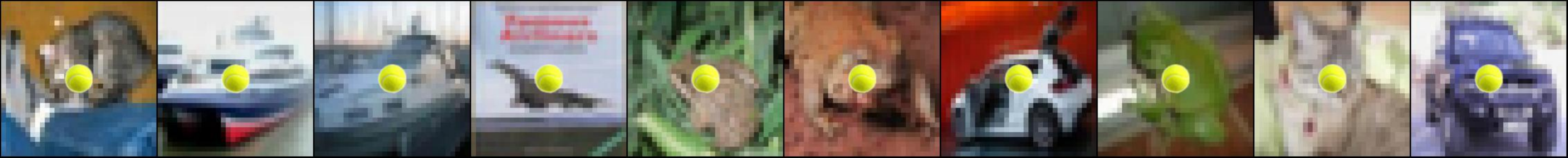}\hfill
        \caption{Example CIFAR-10 images after applying the tennis ball intervention.}
    \end{subfigure}\hfill
    \vspace{1em}
    \begin{subfigure}[c]{\linewidth}
        \includegraphics[width=\linewidth]{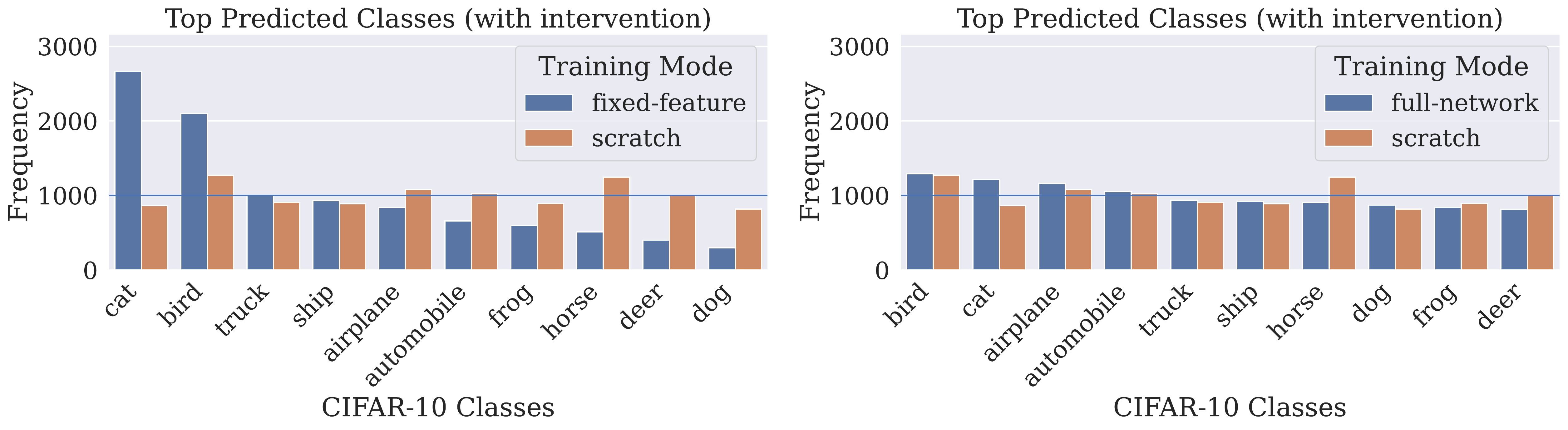}\hfill
        \caption{Distribution of CIFAR-10 model predictions when trained from
        scratch and when transferred from the biased source model. We consider fixed-feature fine-tuning (left) and full-feature fine-tuning (right). The from-scratch models are not affected by the tennis ball
        intervention, while the ones learned via transfer have highly skewed
        output distributions.}
    \end{subfigure}
    \caption{\textbf{The ``tennis ball'' bias.} \textbf{(a-b)} A pre-trained
    ImageNet model is more likely to predict ``tennis ball'' whenever a circular
    yellow shape is in the image. \textbf{(c-d)} This bias transfers to CIFAR-10
    in both fixed-feature and full network transfer settings. Indeed, if we
    overlay a chain-like pattern on all CIFAR-10 test set images as shown above,
    the model predictions skew towards a specific class. This does not happen if
    the CIFAR-10 model was trained from \textit{scratch }instead (orange).} 
    \label{fig:tennisball-bias-results}
\end{figure}

\paragraph{Identifying ImageNet biases.} To identify ImageNet biases, we focus on features that are (a) associated with an ImageNet class and (b) easy to overlay on an image. For example, we used a ``circular yellow shape'' feature is predictive for the class ``tennis ball.'' To verify that these features indeed bias the ImageNet model, we consider a simple counterfactual experiment: we overlay the features on all the ImageNet images and monitor the shift in the model output distribution. As expected, both ``circular yellow shape'' and ``chain-like pattern'' are strong predictive features for the classes ``tennis ball'' and ``chainlink fence''---see Figures~\ref{fig:imagenet-bias-chainlink} and \ref{fig:imagenet-bias-tennisball}. 
These naturally occurring ImageNet biases are thus suitable
for studying the transfer of biases that exist in the wild.

\paragraph{ImageNet-biases transfer to target tasks.} Now, what happens if we
fine-tune a pre-trained ImageNet model (which has these 
biases) on a target dataset such as CIFAR-10? These biases turn out to
persist in the resulting model even though CIFAR-10 does not
contain them (as CIFAR-10 does not contain these classes). To demonstrate this phenomenon, we overlay the associated feature for
both the ``tennis ball'' and ``chainlink fence'' ImageNet classes on the CIFAR-10 test
set. We then evaluate 1) a model fine-tuned on a standard pre-trained
ImageNet model, and 2) a model trained from scratch on the CIFAR-10 dataset.
As Figures~\ref{fig:chainlink-bias-results} and \ref{fig:tennisball-bias-results}
demonstrate, the fine-tuned models (both fixed-feature and full-network) 
are sensitive to the overlaid ImageNet biases, whereas CIFAR-10 models
trained from scratch are not. 
This is further corroborated by the overall skew of the output class distribution
for the transfer-learned model, compared to an almost uniform output class
distribution of the model trained from scratch. 

    \section{Related Work}
    \label{sec:related}
    
\paragraph{Transfer learning.}

Transfer learning has been used in applications ranging from autonomous driving \cite{kim2017end, du2019self}, radiology \cite{wang2017chestx, ke2021chextransfer} to satellite image analysis~\cite{xie2016transfer, wang2019crop}. In particular, fine-tuning pre-trained ImageNet models has increasingly become standard practice to improve the performance of various image classification tasks~\cite{kornblith2019better, salman2020adversarially, utrera2020adversarially}, even on domains substantially different from ImageNet, such as medical imaging \citep{mormont2018comparison, ke2021chextransfer}. Transfer learning from ImageNet is also widely used for most object detection and semantic segmentation tasks \citep{ren2015faster, dai2016r, girshick2014rich, chen2017deeplab}. More recently, even larger vision \cite{radford2021learning, sun2017revisiting} and language models~\cite{brown2020language}---often trained on proprietary datasets---have acted as backbones for downstream tasks. With this widespread usage of pre-trained models, it is important to understand whether any limitation of these models would affect downstream tasks, which is what we focus on in this work.


\paragraph{Backdoor attacks.}
In a backdoor attack~\cite{gu2017badnets, evtimov2018robust, turner2019label}, an adversary maliciously injects a trigger into the source dataset which can be activated during inference. This type of attack can be especially hard to detect, since the model performs well in the absence of the trigger~\cite{gu2017badnets}. Indeed, there exists a long line of work on injecting malicious training examples, known as data poisoning attacks \cite{biggio2012poisoning,xiao2012adversarial,newell2014practicality,mei2015using,steinhardt2017certified}. \citet{gu2017badnets} planted backdoors in a stop-sign detection dataset, and found that fine-tuned stop-sign detection models were still sensitive to this trigger. 


    \section{Conclusion}
    \label{sec:conclusion}
In this work we demonstrated 
that biases in pre-trained models tend to remain present 
even after fine-tuning these models on downstream target tasks. 
Crucially, these biases can persist even when the target dataset 
used for fine-tuning did not contain such biases. 
These findings are of particular concern as researchers and practitioners increasingly leverage 
public pre-trained source models, which are likely to contain undocumented biases.
We thus encourage further investigation of the full machine learning
pipeline---even parts that are seemingly unimportant---for potential sources of 
bias.

    \section{Acknowledgements}
    Work supported in part by the NSF grants CCF-1553428 and CNS-1815221, and Open Philanthropy. This material is based upon work supported by the Defense Advanced Research Projects Agency (DARPA) under Contract No. HR001120C0015.

Research was sponsored by the United States Air Force Research Laboratory and the United States Air Force Artificial Intelligence Accelerator and was accomplished under Cooperative Agreement Number FA8750-19-2-1000. The views and conclusions contained in this document are those of the authors and should not be interpreted as representing the official policies, either expressed or implied, of the United States Air Force or the U.S. Government. The U.S. Government is authorized to reproduce and distribute reprints for Government purposes notwithstanding any copyright notation herein.

Work partially done on the MIT Supercloud compute cluster~\citep{reuther2018interactive}.

    \clearpage
    \printbibliography

@inproceedings{barbu2019objectnet,
 author = {Barbu, Andrei and Mayo, David and Alverio, Julian and Luo, William and Wang, Christopher and Gutfreund, Dan and Tenenbaum, Josh and Katz, Boris},
 booktitle = {Neural Information Processing Systems (NeurIPS)},
 title = {ObjectNet: A large-scale bias-controlled dataset for pushing the limits of object recognition models},
 year = {2019}
}

@inproceedings{berg2014birdsnap,
 author = {Berg, Thomas and Liu, Jiongxin and Woo Lee, Seung and Alexander, Michelle L and Jacobs, David W and Belhumeur, Peter N},
 booktitle = {Proceedings of the IEEE Conference on Computer Vision and Pattern Recognition},
 title = {Birdsnap: Large-scale fine-grained visual categorization of birds},
 year = {2014}
}

@inproceedings{biggio2012poisoning,
 author = {Biggio, Battista and Nelson, Blaine and Laskov, Pavel},
 booktitle = {International Conference on Machine Learning},
 title = {Poisoning attacks against support vector machines},
 year = {2012}
}

@inproceedings{bossard2014food,
 author = {Bossard, Lukas and Guillaumin, Matthieu and Van Gool, Luc},
 booktitle = {European conference on computer vision},
 title = {Food-101--mining discriminative components with random forests},
 year = {2014}
}

@article{brown2020language,
 author = {Brown, Tom B and Mann, Benjamin and Ryder, Nick and Subbiah, Melanie and Kaplan, Jared and Dhariwal, Prafulla and Neelakantan, Arvind and Shyam, Pranav and Sastry, Girish and Askell, Amanda and others},
 journal = {arXiv preprint arXiv:2005.14165},
 title = {Language models are few-shot learners},
 year = {2020}
}

@inproceedings{buolamwini2018gender,
 author = {Buolamwini, Joy and Gebru, Timnit},
 booktitle = {Conference on fairness, accountability and transparency (FAccT)},
 title = {Gender shades: Intersectional accuracy disparities in commercial gender classification},
 year = {2018}
}

@article{chen2017deeplab,
 author = {Chen, Liang-Chieh and Papandreou, George and Kokkinos, Iasonas and Murphy, Kevin and Yuille, Alan L},
 journal = {IEEE transactions on pattern analysis and machine intelligence},
 title = {Deeplab: Semantic image segmentation with deep convolutional nets, atrous convolution, and fully connected crfs},
 year = {2017}
}

@inproceedings{dai2016r,
 author = {Dai, Jifeng and Li, Yi and He, Kaiming and Sun, Jian},
 booktitle = {Advances in neural information processing systems (NeurIPS)},
 title = {R-fcn: Object detection via region-based fully convolutional networks},
 year = {2016}
}

@inproceedings{deng2009imagenet,
 author = {Deng, Jia and Dong, Wei and Socher, Richard and Li, Li-Jia and Li, Kai and Fei-Fei, Li},
 booktitle = {Computer Vision and Pattern Recognition (CVPR)},
 title = {Imagenet: A large-scale hierarchical image database},
 year = {2009}
}

@article{du2019self,
 author = {Du, Shuyang and Guo, Haoli and Simpson, Andrew},
 journal = {arXiv preprint arXiv:1912.05440},
 title = {Self-driving car steering angle prediction based on image recognition},
 year = {2019}
}

@inproceedings{evtimov2018robust,
 author = {Evtimov, Ivan and Eykholt, Kevin and Fernandes, Earlence
and Kohno, Tadayoshi and Li, Bo and Prakash, Atul and
Rahmati, Amir and Song, Dawn},
 booktitle = {Conference on Computer Vision and Pattern Recognition
(CVPR)},
 title = {Robust Physical-World Attacks on Machine Learning Models},
 year = {2018}
}

@inproceedings{fei2004learning,
 author = {Fei-Fei, Li and Fergus, Rob and Perona, Pietro},
 booktitle = {2004 conference on computer vision and pattern recognition workshop},
 organization = {IEEE},
 pages = {178--178},
 title = {Learning generative visual models from few training examples: An incremental bayesian approach tested on 101 object categories},
 year = {2004}
}

@inproceedings{girshick2014rich,
 author = {Girshick, Ross and Donahue, Jeff and Darrell, Trevor and Malik, Jitendra},
 booktitle = {computer vision and pattern recognition (CVPR)},
 pages = {580--587},
 title = {Rich feature hierarchies for accurate object detection and semantic segmentation},
 year = {2014}
}

@article{griffin2007caltech,
 author = {Griffin, Gregory and Holub, Alex and Perona, Pietro},
 publisher = {California Institute of Technology},
 title = {Caltech-256 object category dataset},
 year = {2007}
}

@article{gu2017badnets,
 author = {Gu, Tianyu and Dolan-Gavitt, Brendan and Garg, Siddharth},
 journal = {arXiv preprint arXiv:1708.06733},
 title = {Badnets: Identifying Vulnerabilities in the Machine Learning Model Supply Chain},
 year = {2017}
}

@inproceedings{ke2021chextransfer,
 author = {Ke, Alexander and Ellsworth, William and Banerjee, Oishi and Ng, Andrew Y and Rajpurkar, Pranav},
 booktitle = {Proceedings of the Conference on Health, Inference, and Learning},
 pages = {116--124},
 title = {CheXtransfer: performance and parameter efficiency of ImageNet models for chest X-Ray interpretation},
 year = {2021}
}

@inproceedings{kim2017end,
 author = {Kim, Jiman and Park, Chanjong},
 booktitle = {Proceedings of the IEEE conference on computer vision and pattern recognition workshops},
 pages = {30--38},
 title = {End-to-end ego lane estimation based on sequential transfer learning for self-driving cars},
 year = {2017}
}

@inproceedings{kornblith2019better,
 author = {Kornblith, Simon and Shlens, Jonathon and Le, Quoc V},
 booktitle = {computer vision and pattern recognition (CVPR)},
 title = {Do better imagenet models transfer better?},
 year = {2019}
}

@article{krause2013collecting,
 author = {Krause, Jonathan and Deng, Jia and Stark, Michael and Fei-Fei, Li},
 title = {Collecting a large-scale dataset of fine-grained cars},
 year = {2013}
}

@inproceedings{krizhevsky2009learning,
 author = {Krizhevsky, Alex},
 booktitle = {Technical report},
 title = {Learning Multiple Layers of Features from Tiny Images},
 year = {2009}
}

@inproceedings{leclerc20213db,
 author = {Leclerc, Guillaume and Salman, Hadi and Ilyas, Andrew and Vemprala, Sai and Engstrom, Logan and Vineet, Vibhav and Xiao, Kai and Zhang, Pengchuan and Santurkar, Shibani and Yang, Greg and others},
 booktitle = {arXiv preprint arXiv:2106.03805},
 title = {3DB: A Framework for Debugging Computer Vision Models},
 year = {2021}
}

@misc{leclerc2022ffcv,
 author = {Guillaume Leclerc and Andrew Ilyas and Logan Engstrom and Sung Min Park and Hadi Salman and Aleksander Madry},
 howpublished = {\url{https://github.com/libffcv/ffcv/}},
 title = {ffcv},
 year = {2022}
}

@inproceedings{lin2014microsoft,
 author = {Lin, Tsung-Yi and Maire, Michael and Belongie, Serge and Hays, James and Perona, Pietro and Ramanan, Deva and Doll{\'a}r, Piotr and Zitnick, C Lawrence},
 booktitle = {European conference on computer vision (ECCV)},
 title = {Microsoft coco: Common objects in context},
 year = {2014}
}

@inproceedings{mahajan2018exploring,
 author = {Mahajan, Dhruv and Girshick, Ross and Ramanathan, Vignesh and He, Kaiming and Paluri, Manohar and Li, Yixuan and Bharambe, Ashwin and van der Maaten, Laurens},
 booktitle = {Proceedings of the European Conference on Computer Vision (ECCV)},
 title = {Exploring the limits of weakly supervised pretraining},
 year = {2018}
}

@article{maji2013fine,
 author = {Maji, Subhransu and Rahtu, Esa and Kannala, Juho and Blaschko, Matthew and Vedaldi, Andrea},
 journal = {arXiv preprint arXiv:1306.5151},
 title = {Fine-grained visual classification of aircraft},
 year = {2013}
}

@inproceedings{mei2015using,
 author = {Mei, Shike and Zhu, Xiaojin},
 booktitle = {AAAI},
 pages = {2871--2877},
 title = {Using Machine Teaching to Identify Optimal Training-Set Attacks on Machine Learners.},
 year = {2015}
}

@inproceedings{mormont2018comparison,
 author = {Mormont, Romain and Geurts, Pierre and Mar{\'e}e, Rapha{\"e}l},
 booktitle = {Proceedings of the IEEE Conference on Computer Vision and Pattern Recognition Workshops},
 title = {Comparison of deep transfer learning strategies for digital pathology},
 year = {2018}
}

@inproceedings{newell2014practicality,
 author = {Newell, Andrew and Potharaju, Rahul and Xiang, Luojie and Nita-Rotaru, Cristina},
 booktitle = {Proceedings of the 2014 Workshop on Artificial Intelligent and Security Workshop},
 organization = {ACM},
 pages = {83--93},
 title = {On the Practicality of Integrity Attacks on Document-Level Sentiment Analysis},
 year = {2014}
}

@inproceedings{nilsback2008automated,
 author = {Nilsback, Maria-Elena and Zisserman, Andrew},
 booktitle = {2008 Sixth Indian Conference on Computer Vision, Graphics \& Image Processing},
 title = {Automated flower classification over a large number of classes},
 year = {2008}
}

@inproceedings{parkhi2012cats,
 author = {Parkhi, Omkar M and Vedaldi, Andrea and Zisserman, Andrew and Jawahar, CV},
 booktitle = {2012 IEEE conference on computer vision and pattern recognition},
 organization = {IEEE},
 pages = {3498--3505},
 title = {Cats and dogs},
 year = {2012}
}

@inproceedings{radford2021learning,
 author = {Radford, Alec and Kim, Jong Wook and Hallacy, Chris and Ramesh, Aditya and Goh, Gabriel and Agarwal, Sandhini and Sastry, Girish and Askell, Amanda and Mishkin, Pamela and Clark, Jack and others},
 booktitle = {arXiv preprint arXiv:2103.00020},
 title = {Learning transferable visual models from natural language supervision},
 year = {2021}
}

@inproceedings{ren2015faster,
 author = {Ren, Shaoqing and He, Kaiming and Girshick, Ross and Sun, Jian},
 booktitle = {Advances in neural information processing systems (NeurIPS)},
 title = {Faster r-cnn: Towards real-time object detection with region proposal networks},
 year = {2015}
}

@inproceedings{reuther2018interactive,
 author = {Reuther, Albert and Kepner, Jeremy and Byun, Chansup and Samsi, Siddharth and Arcand, William and Bestor, David and Bergeron, Bill and Gadepally, Vijay and Houle, Michael and Hubbell, Matthew and Jones, Michael and Klein, Anna and Milechin, Lauren and Mullen, Julia and Prout, Andrew and Rosa, Antonio and Yee, Charles and Michaleas, Peter},
 booktitle = {2018 IEEE High Performance extreme Computing Conference (HPEC)},
 organization = {IEEE},
 pages = {1--6},
 title = {Interactive supercomputing on 40,000 cores for machine learning and data analysis},
 year = {2018}
}

@inproceedings{russakovsky2015imagenet,
 author = {Olga Russakovsky and Jia Deng and Hao Su and Jonathan
Krause and Sanjeev Satheesh and Sean Ma and Zhiheng Huang
and Andrej Karpathy and Aditya Khosla and Michael Bernstein
and Alexander C. Berg and Li Fei-Fei},
 booktitle = {International Journal of Computer Vision (IJCV)},
 title = {{ImageNet Large Scale Visual Recognition Challenge}},
 year = {2015}
}

@inproceedings{salman2020adversarially,
 author = {Salman, Hadi and Ilyas, Andrew and Engstrom, Logan and Kapoor, Ashish and Madry, Aleksander},
 booktitle = {Advances in Neural Information Processing Systems (NeurIPS)},
 title = {Do Adversarially Robust ImageNet Models Transfer Better?},
 year = {2020}
}

@inproceedings{steinhardt2017certified,
 author = {Steinhardt, Jacob and Koh, Pang Wei W and Liang, Percy S},
 booktitle = {NIPS},
 title = {Certified Defenses for Data Poisoning Attacks},
 year = {2017}
}

@inproceedings{sun2017revisiting,
 author = {Sun, Chen and Shrivastava, Abhinav and Singh, Saurabh and Gupta, Abhinav},
 booktitle = {Proceedings of the IEEE international conference on computer vision},
 title = {Revisiting unreasonable effectiveness of data in deep learning era},
 year = {2017}
}

@article{terhorst2021comprehensive,
 author = {Terh{\"o}rst, Philipp and Kolf, Jan Niklas and Huber, Marco and Kirchbuchner, Florian and Damer, Naser and Morales, Aythami and Fierrez, Julian and Kuijper, Arjan},
 journal = {arXiv preprint arXiv:2103.01592},
 title = {A comprehensive study on face recognition biases beyond demographics},
 year = {2021}
}

@inproceedings{turner2019label,
 author = {Turner, Alexander and Tsipras, Dimitris and Madry, Aleksander},
 title = {Label-Consistent Backdoor Attacks},
 year = {2019}
}

@inproceedings{utrera2020adversarially,
 author = {Francisco Utrera and Evan Kravitz and N. Benjamin Erichson and Rajiv Khanna and Michael W. Mahoney},
 booktitle = {ArXiv preprint arXiv:2007.05869},
 title = {Adversarially-Trained Deep Nets Transfer Better},
 year = {2020}
}

@inproceedings{wang2017chestx,
 author = {Wang, Xiaosong and Peng, Yifan and Lu, Le and Lu, Zhiyong and
Bagheri, Mohammadhadi and Summers, Ronald M},
 booktitle = {Proceedings of the IEEE conference on computer vision and
pattern recognition (CVPR)},
 title = {Chestx-ray8: Hospital-scale chest x-ray database and benchmarks
on weakly-supervised classification and localization of common thorax
diseases},
 year = {2017}
}

@article{wang2019crop,
 author = {Wang, Sherrie and Azzari, George and Lobell, David B},
 journal = {Remote sensing of environment},
 pages = {303--317},
 publisher = {Elsevier},
 title = {Crop type mapping without field-level labels: Random forest transfer and unsupervised clustering techniques},
 volume = {222},
 year = {2019}
}

@inproceedings{xiao2010sun,
 author = {Xiao, Jianxiong and Hays, James and Ehinger, Krista A and Oliva, Aude and Torralba, Antonio},
 booktitle = {Computer Vision and Pattern Recognition (CVPR)},
 title = {Sun database: Large-scale scene recognition from abbey to zoo},
 year = {2010}
}

@inproceedings{xiao2012adversarial,
 author = {Xiao, Han and Xiao, Huang and Eckert, Claudia},
 booktitle = {European Conference on Artificial Intelligence (ECAI)},
 title = {Adversarial Label Flips Attack on Support Vector Machines.},
 year = {2012}
}

@article{xiao2020noise,
 author = {Xiao, Kai and Engstrom, Logan and Ilyas, Andrew and Madry, Aleksander},
 journal = {arXiv preprint arXiv:2006.09994},
 title = {Noise or signal: The role of image backgrounds in object recognition},
 year = {2020}
}

@inproceedings{xie2016transfer,
 author = {Xie, Michael and Jean, Neal and Burke, Marshall and Lobell, David and Ermon, Stefano},
 booktitle = {Thirtieth AAAI Conference on Artificial Intelligence},
 title = {Transfer learning from deep features for remote sensing and poverty mapping},
 year = {2016}
}

@inproceedings{zhu2017object,
 author = {Zhuotun Zhu and Lingxi Xie and Alan Yuille},
 booktitle = {International Joint Conference on Artificial Intelligence},
 title = {Object Recognition without and without Objects},
 year = {2017}
}

@inproceedings{geirhos2018imagenettrained,
title={ImageNet-trained {CNN}s are biased towards texture; increasing shape bias improves accuracy and robustness.},
author={Robert Geirhos and Patricia Rubisch and Claudio Michaelis and Matthias Bethge and Felix A. Wichmann and Wieland Brendel},
booktitle={International Conference on Learning Representations (ICLR)},
year={2019},
}

@inproceedings{liu2015faceattributes,
  title = {Deep Learning Face Attributes in the Wild},
  author = {Liu, Ziwei and Luo, Ping and Wang, Xiaogang and Tang, Xiaoou},
  booktitle = {International Conference on Computer Vision
    (ICCV)},
  year = {2015}
}

@inproceedings{singh2020don,
  title={Don't judge an object by its context: learning to overcome contextual bias},
  author={Singh, Krishna Kumar and Mahajan, Dhruv and Grauman, Kristen and Lee, Yong Jae and Feiszli, Matt and Ghadiyaram, Deepti},
  booktitle={Proceedings of the IEEE/CVF Conference on Computer Vision and Pattern Recognition},
  pages={11070--11078},
  year={2020}
}
    \clearpage

    \appendix
    \section{Experimental Setup}
    \label{app:setup}
    
\subsection{ImageNet Models}
\label{app:models}
In this paper, we train a number of ImageNet models and transfer them to various datasets in Sections~\ref{sec:main-exp} and \ref{sec:ultra_realistic}. We mainly use the ResNet-18 architecture all over the paper. However, we study bias transfers using various architectures in Appendix~\ref{app:vary-arch}. We use PyTorch's official implementation for these architectures, which can be found here \url{https://pytorch.org/vision/stable/models.html}. 

\paragraph{Training details.}
We train our ImageNet models from scratch using SGD by minimizing the standard cross-entropy loss. We train for 16 epochs using a Cyclic learning rate schedule with an initial learning rate of $0.5$ and learning rate peak epoch of $2$. We use momentum of $0.9$, batch size of 1024, and weight decay of $5e^{-4}$. We use standard data-augmentation: \textit{RandomResizedCrop} and 
\textit{RandomHorizontalFlip} during training, and \textit{RandomResizedCrop} during testing. Our implementation and configuration files are available in the attached code.

\subsection{Transfer details from ImageNet to downstream image classification tasks}
\label{app:transfer-to-small-datasets}

\paragraph{Transfer datasets.}
\label{app:classification-datasets}

We use the  image classification tasks that are used in \citep{salman2020adversarially,kornblith2019better}, which have various sizes and number of classes. When evaluating the performance of models on each of these datasets, we report the Top-1 accuracy for balanced datasets and the Mean Per-Class accuracy for the unbalanced datasets. See Table~\ref{table:datasets} for the details of these datasets. For each dataset, we consider two transfer learning settings: \textit{fixed-feature} and \textit{full-network} transfer learning which we describe below.

\begin{table}[!h]
\caption{Image classification benchmarks used in this paper. Accuracy metric is the metric we report for each of the dataset across the paper. Some datasets are imbalanced, so we report Mean Per-Class accuracy for those. For the rest, we report Top-1 accuracy.}
\label{table:datasets}
    \begin{center}
        \begin{small}
        \begin{tabular}{@{}lccc@{}}
        \toprule
        \textbf{Dataset}           & \textbf{Size (Train/Test)} &\textbf{Classes} &  \textbf{Accuracy Metric}\\ \midrule
        Birdsnap \citep{berg2014birdsnap}                       & 32,677/8,171 & 500 & Top-1      \\ 
        Caltech-101 \citep{fei2004learning} &  3,030/5,647 &101 & Mean Per-Class\\
        Caltech-256 \citep{griffin2007caltech}  & 15,420/15,187 & 257& Mean Per-Class\\ 
        CIFAR-10 \citep{krizhevsky2009learning}                       & 50,000/10,000  & 10 & Top-1   \\ 
        CIFAR-100 \citep{krizhevsky2009learning}                      & 50,000/10,000  & 100& Top-1     \\ 
        FGVC Aircraft \citep{maji2013fine}                 & 6,667/3,333 & 100  & Mean Per-Class       \\ 
        Food-101 \citep{bossard2014food}                       & 75,750/25,250 & 101 & Top-1     \\ 
        Oxford 102 Flowers \citep{nilsback2008automated}           & 2,040/6,149 & 102   & Mean Per-Class         \\
        Oxford-IIIT Pets \cite{parkhi2012cats} & 3,680/3,669 &37 &  Mean Per-Class\\
        SUN397 \citep{xiao2010sun}                        & 19,850/19,850  & 397 & Top-1     \\ 
        Stanford Cars \citep{krause2013collecting}                  & 8,144/8,041  & 196 & Top-1      \\ 
        \bottomrule
    \end{tabular}
    \end{small}
    \end{center}
\end{table}

\paragraph{Fixed-feature transfer.}
\label{app:logistic-regression-params}
For this setting, we \textit{freeze} the layers of the ImageNet source model\footnote{We do not freeze the batch norm statistics, but only the weights of the model similar to~\citet{salman2020adversarially}.}, except for the last layer, which we replace with a random initialized linear layer whose output matches the number of classes in the transfer dataset.
We now train only this new layer for using SGD, with a batch size of 1024 using cyclic learning rate. For more details and hyperparameter for each dataset, please see config files in the attached code. 

\paragraph{Full-network transfer.}
\label{app:finetuning-params}
For this setting, we \textit{do not freeze} any of the layers of the ImageNet source model, and all the model weights are updated. We follow the exact same hyperparameters as the fixed-feature setting.

\subsection{Compute and training time}
Throughout the paper, we use the FFCV data-loading library to train models fast ~\cite{leclerc2022ffcv}. Using FFCV, we can train an ImageNet model, for example, in around 1 hr only on a single V100 GPU. Our experiments were conducted on a GPU cluster containing A100 and V100 GPUs.

\subsection{Varying architectures}
\label{app:vary-arch}
In this section, we study whether bias transfers when applying transfer learning using various architectures. We conduct the basic experiment of Section~\ref{sec:main-exp} on several standard architectures from the PyTorch's Torchvision\footnote{These models can be found here \url{https://pytorch.org/vision/stable/models.html}}. 

As in Section~\ref{sec:main-exp}, we train two versions of each architecture: one on a clean ImageNet dataset, and another on a modified ImageNet dataset containing a backdoor.  We use the same hyperparameters as the rest of the paper, except for the batch size, which we set to 512 instead of 1024. The reason we lower the batch size is to fit these models in memory on a single A100 GPU. 

Now, we transfer each of these models to a clean CIFAR-10 dataset, and test if the backdoor attack transfers. Similar to the results of the main paper, we notice that backdoor attack indeed transfers in the fixed-feature setting. We note however that for the full-network setting, all architectures other than ResNet-18 (which we use in the rest of the paper) seem to be more robust to the backdoor attack.

\begin{figure}[!htbp]
    \centering
    \begin{subfigure}[t]{0.88\linewidth}
        \includegraphics[width=\linewidth]{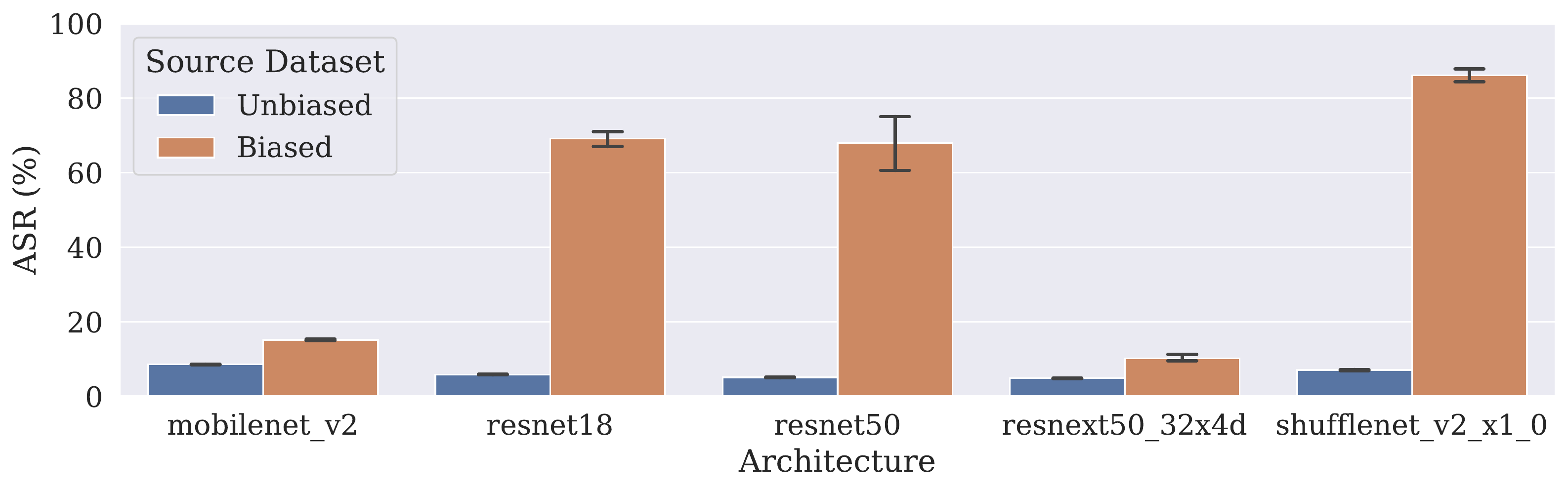}
        \caption{CIFAR-10 Fixed-feature}
        \label{subfig:vary_arch_fixed-feature}
    \end{subfigure}\hfill
    \begin{subfigure}[t]{0.88\linewidth}
        \includegraphics[width=\linewidth]{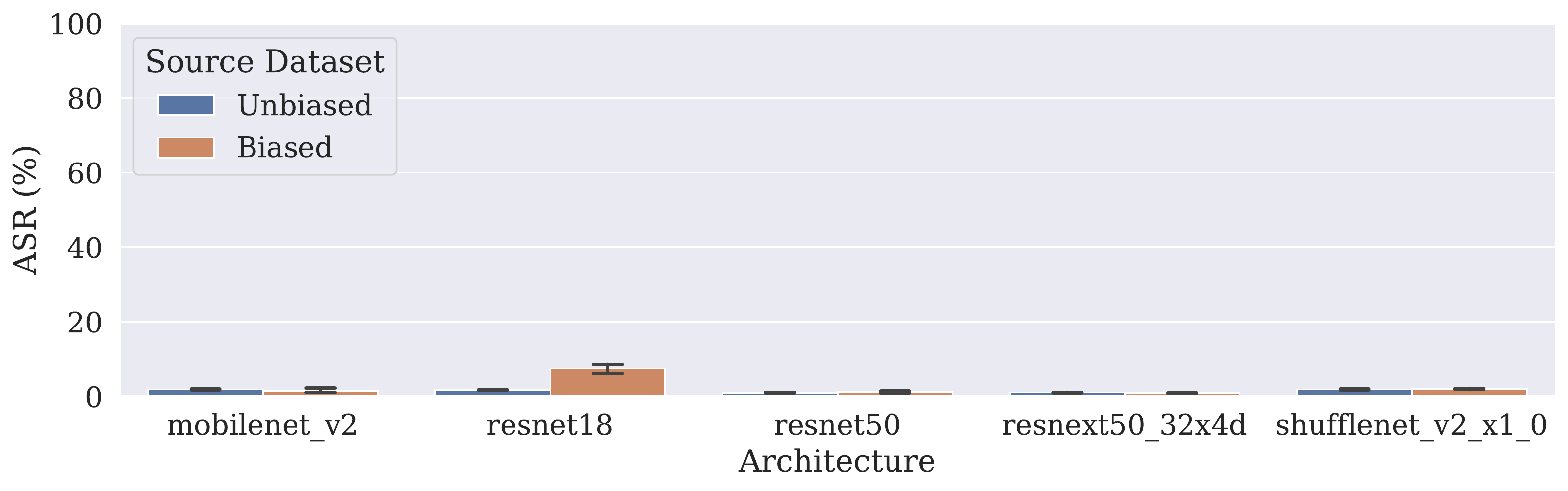}
        \caption{CIFAR-10 Full-network}
        \label{subfig:vary_arch_full-network}
    \end{subfigure}
    \caption{Backdoor attack (bias) consistently transfers in the fixed-feature setting across various architectures. However, this happens to a lesser degree in the full-network transfer setting.}
    \label{fig:vary_arch}
\end{figure}

\subsection{MS-COCO}
\label{app:ms_coco_setup}
In this section, we provide experimental details for the experiment on MS-COCO in Section~\ref{sec:mscoco}. We consider the binary task of predicting cats from dogs, where there is a strong correlation between dogs and the presence of people.

\paragraph{Dataset construction.} We create two source datasets which are described in Table~\ref{apptable:mscoco_ds}.

\begin{table}[h!]
    \centering
    \caption{The synthetic datasets we create from MS-COCO for the experiment in Section~\ref{sec:mscoco}.}
    \label{apptable:mscoco_ds}
    \begin{tabular}[c]{c|c c c c}
        \toprule
        & \multicolumn{2}{c}{\textit{Class: Cat}} &  \multicolumn{2}{c}{\textit{Class: Dog}}\\
        Dataset & With People & Without People & With People & Without People\\
        \midrule
        Non-Spurious & 0 & 1000 & 0 & 100\\
        Spurious & 1000 & 4000 & 4000 & 1000\\
        \bottomrule
    \end{tabular}
\end{table}
We then fine-tune models trained on the above source datasets on new images of cats and dogs without people (485 each). We use the cats and dogs from the MS-COCO test set for evaluation.

\paragraph{Experimental details.} We train a ResNet-18 with resolution $224 \times 224$. We use SGD with momentum, and a Cyclic learning rate. We use the following hyperparameters shown in Table~\ref{apptable:mscoco-hyperparams}:
\begin{table}[h!]
    \centering
    \caption{Hyperparameters used for training on the MS-COCO dataset.}
    \label{apptable:mscoco-hyperparams}
    \begin{tabular}[c]{r|c c}
        \toprule
        Hyperparameter & Value for pre-training & Value for fine-tuning\\
        \midrule
        Batch Size & 256 & 256\\
        Epochs & 25 & 25\\
        LR & 0.01  & 0.005\\
        Momentum & 0.9 & 0.9\\
        Weight Decay & 0.00005 &  0.00005 \\
        Peak Epoch & 2 & 2 \\
        \bottomrule
    \end{tabular}
\end{table}

\subsection{CelebA}
\label{app:celeba_setup_synth}
In this section, we provide experimental details for the CelebA experiments in Section~\ref{sec:celeba}. Here, the task was to distinguish old from young faces, in the presence of a spurious correlation with gender in the source dataset.

\paragraph{Dataset construction.} We create two source datasets shown in Table~\ref{apptable:celeba_ds}:

\begin{table}[h!]
    \centering
    \caption{The synthetic source datasets we create from CelebA for the experiment in Section~\ref{sec:celeba}.}
    \label{apptable:celeba_ds}
    \begin{tabular}[c]{c|c c c c}
        \toprule
        & \multicolumn{2}{c}{\textit{Class: Young}} &  \multicolumn{2}{c}{\textit{Class: Old}}\\
        Dataset & Male & Female & Male & Female\\
        \midrule
        Non-Spurious & 2500 & 2500 & 2500 & 2500\\
        Spurious & 1000 & 4000 & 4000 & 1000\\
        \bottomrule
    \end{tabular}
\end{table}

Due to imbalances in the spurious dataset, the model trained on this dataset struggles on faces of young males and old females. We then fine-tune the source models on the following target datasets (see Table~\ref{apptable:celeba_ds_target}), the images of which are disjoint from that in the source dataset.

\begin{table}[h!]
    \centering
    \caption{The synthetic target datasets we create from CelebA for the experiment in Section~\ref{sec:celeba}.}
    \label{apptable:celeba_ds_target}
    \begin{tabular}[c]{c|c c c c}
        \toprule
        & \multicolumn{2}{c}{\textit{Class: Young}} &  \multicolumn{2}{c}{\textit{Class: Old}}\\
        Dataset & Male & Female & Male & Female\\
        \midrule
        Only Women & 0 & 5000 & 0 & 5000\\
        80\% Women$\mid$20\% Men & 1000 & 4000 & 1000 & 4000\\
        50\% Women$\mid$50\% Men & 2500 & 2500 & 2500 & 2500\\
        \bottomrule
    \end{tabular}
\end{table}
Due to space constraints, we plotted the results of fixed fine-tuning on Only Women and 80\% Women$\mid$20\% Men in the main paper. Below, we display the results for fixed-feature and full fine-tuning on all 3 target datasets.

\paragraph{Experimental details.} We train a ResNet-18 with resolution $224 \times 224$. We use SGD with momentum, and a cyclic learning rate. We use the following hyperparameters shown in Table~\ref{apptable:celeba-hyperparams}:
\begin{table}[h!]
    \centering
    \caption{Hyperparameters used for training on the CelebA datasets.}
    \label{apptable:celeba-hyperparams}
    \begin{tabular}[c]{cccccc}
        \toprule
        Batch Size & Epochs & LR & Momentum & Weight Decay & Peak Epoch\\
        \midrule
        1024 & 20 & 0.05 & 0.9 & 0.01 & 5 \\
        \bottomrule
    \end{tabular}
\end{table}

\paragraph{Results.}
We find that in both the fixed-feature and full-feature fine-tuning settings, the gender correlation transfers from the source model to the transfer model, even though the target task is itself gender balanced. As the proportion of men and women in the target dataset change, the model is either more sensitive to the presence of women, or more sensitive to the presence of men. In all cases, however, the model transferred from the spurious backbone is more sensitive to gender than a model transferred from the non-spurious backbone.
\begin{figure}[!b]
    \centering
    \begin{subfigure}[t]{0.32\linewidth}
        \centering
        \includegraphics[width=\linewidth]{figures/facial_recognition/celeba_semi_synthetic/celeba_base_accs.pdf}
        \caption{Source Model Accuracy}
    \end{subfigure}\\
    \begin{subfigure}[t]{0.32\linewidth}
        \centering
        \includegraphics[width=\linewidth]{figures/facial_recognition/celeba_semi_synthetic/fixed_finetune_only_women.pdf}
        \caption{Fixed-Feature Fine-tuning. Target Dataset: Only Women}
    \end{subfigure}\hfill
    \begin{subfigure}[t]{0.32\linewidth}
        \centering
        \includegraphics[width=\linewidth]{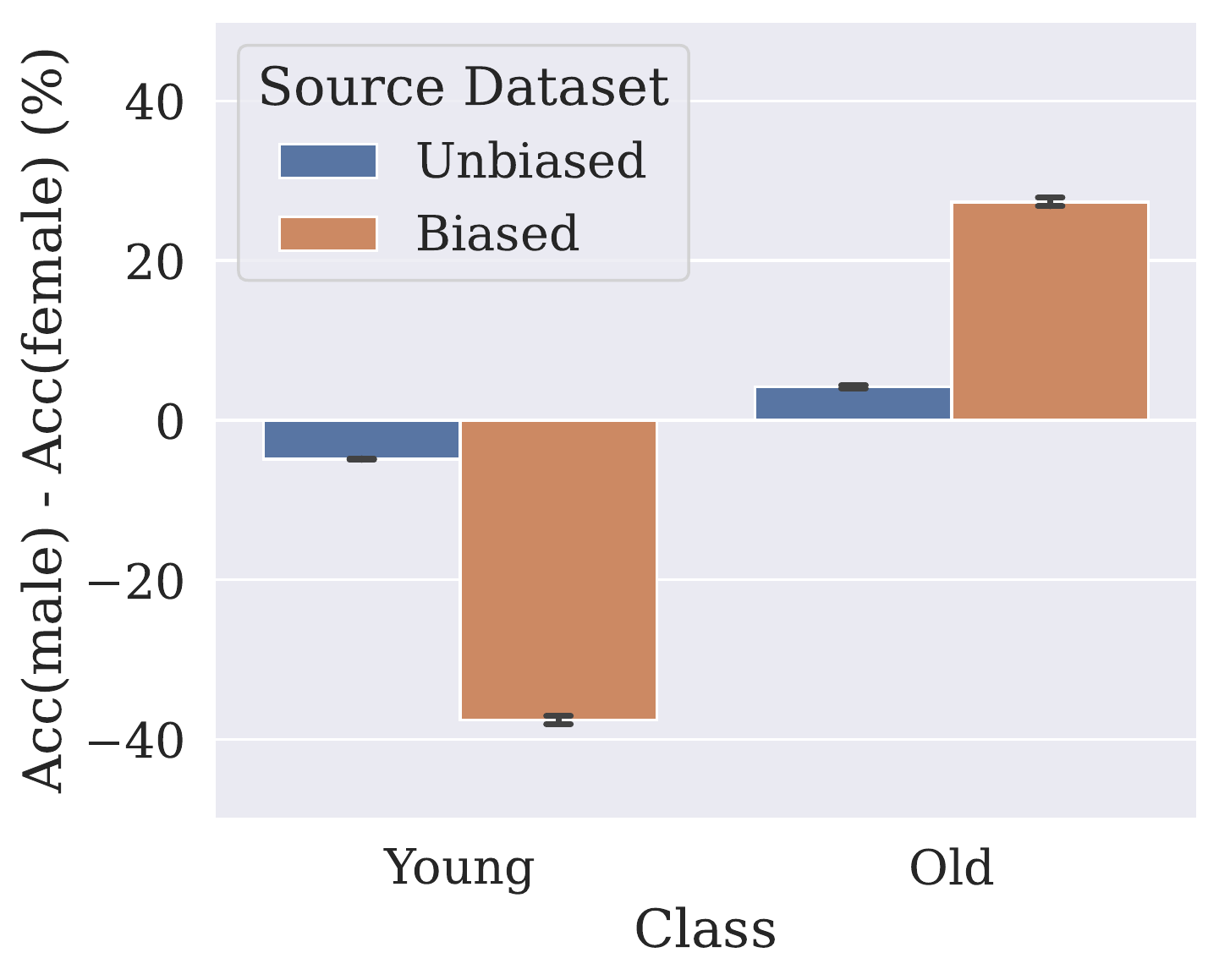}
        \caption{Fixed-Feature Fine-tuning. Target Dataset: 80\% Women, 20\% Men}
    \end{subfigure}\hfill
    \begin{subfigure}[t]{0.32\linewidth}
        \centering
        \includegraphics[width=\linewidth]{figures/facial_recognition/celeba_semi_synthetic/fixed_finetune_eq_women_and_men.pdf}
        \caption{Fixed-Feature Fine-tuning. Target Dataset:  50\% Women, 50\% Men}
    \end{subfigure}\hfill
    \begin{subfigure}[t]{0.32\linewidth}
        \centering
        \includegraphics[width=\linewidth]{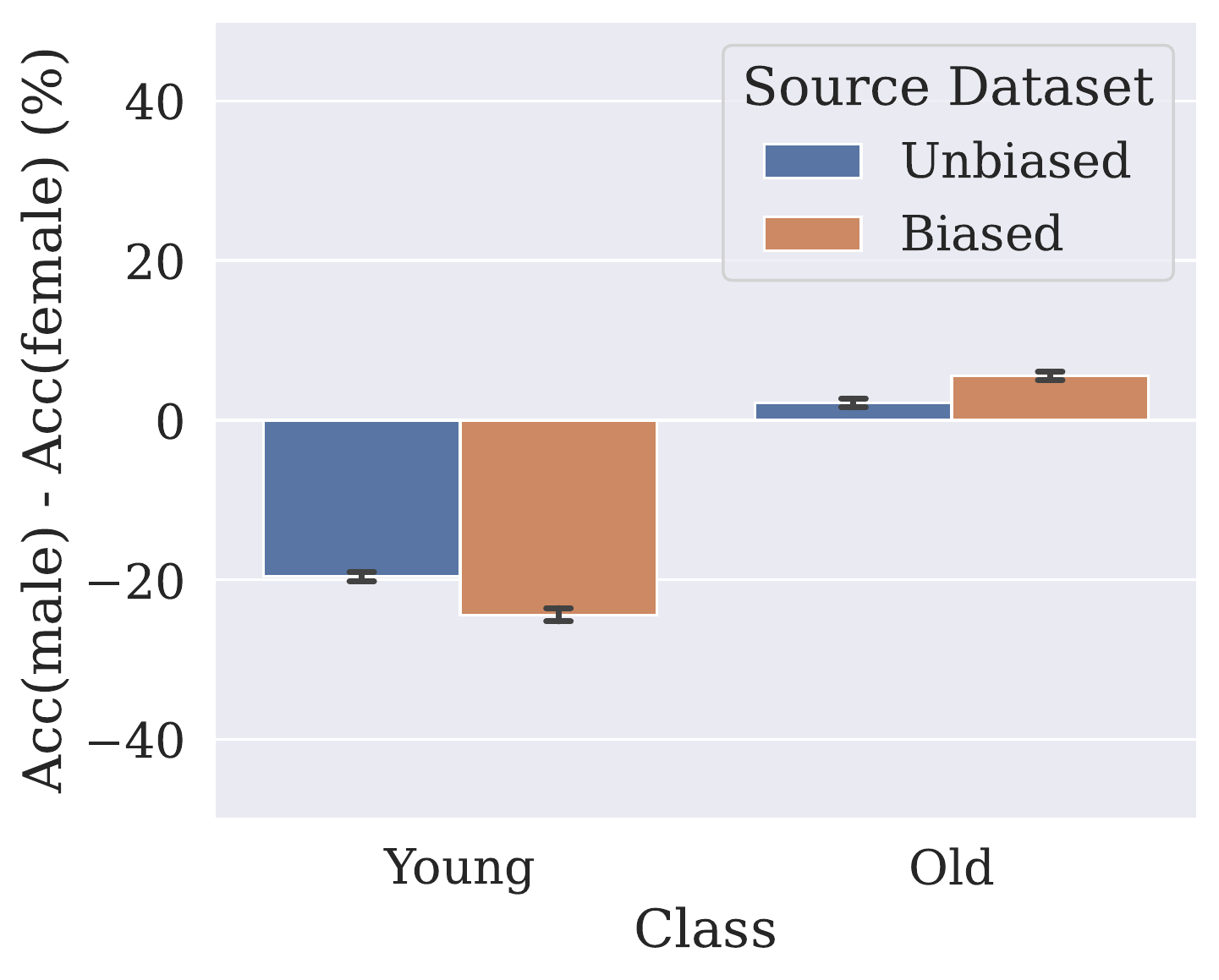}
        \caption{Full-Feature Fine-tuning. Target Dataset: Only Women}
    \end{subfigure}\hfill
    \begin{subfigure}[t]{0.32\linewidth}
        \centering
        \includegraphics[width=\linewidth]{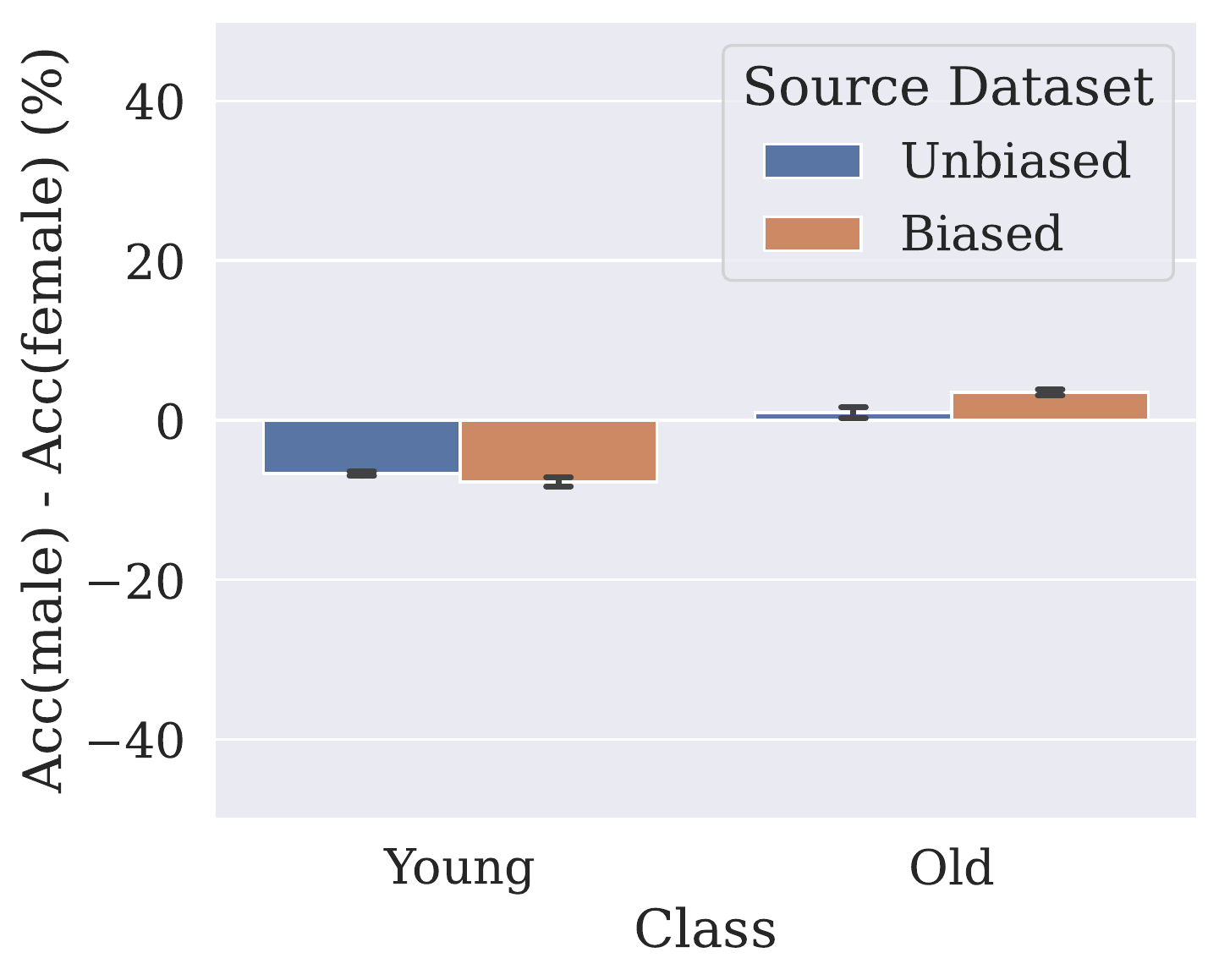}
        \caption{Full-Feature Fine-tuning. Target Dataset: 80\% Women, 20\% Men}
    \end{subfigure}\hfill
    \begin{subfigure}[t]{0.32\linewidth}
        \centering
        \includegraphics[width=\linewidth]{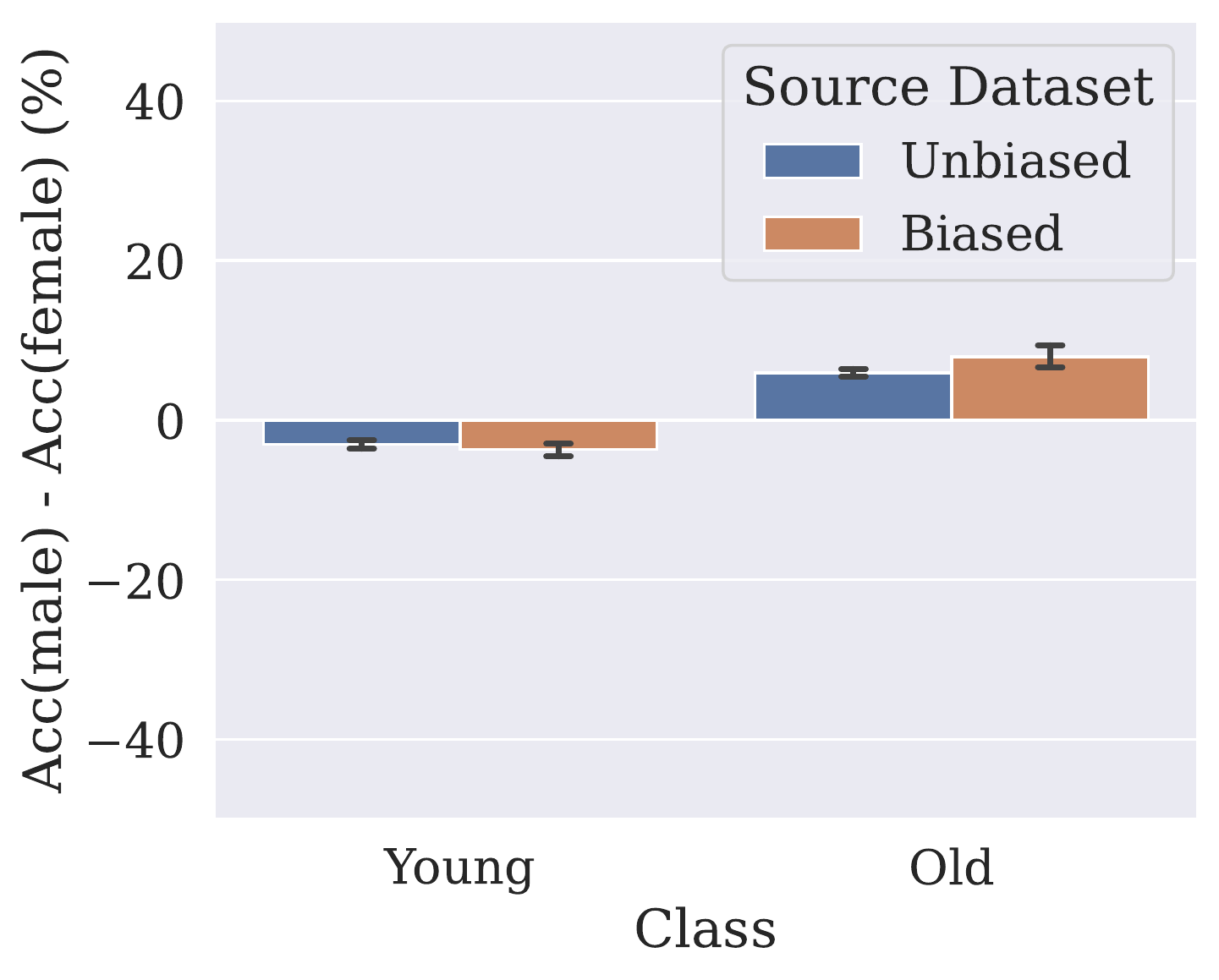}
        \caption{Full-Feature Fine-tuning. Target Dataset:  50\% Women, 50\% Men}
    \end{subfigure}
    \caption{
        \textbf{CelebA Experiment.}We consider transfer from a source dataset that spuriously correlate age with gender --- such that old men and young women are overrepresented. We plot the difference in accuracies between male and female examples, and find that the model transferred from a spurious backbone is sensitive to gender, even though the target dataset was itself gender balanced. }
        \label{app_fig:celeba}
\end{figure}

    \clearpage
    \section{ImageNet Biases}
    \label{app:imagenet-biases}
    \subsection{Chainlink fence bias.}
\begin{figure}[!htbp]
    \begin{subfigure}{\linewidth}
        \includegraphics[width=\linewidth,trim={0 0 0 0},clip]{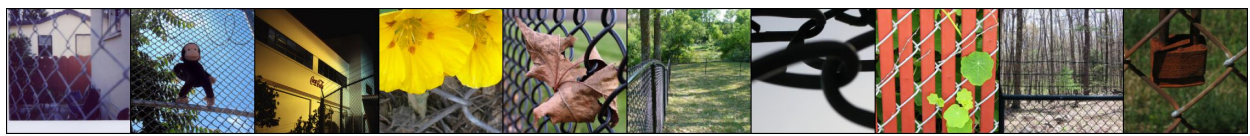}
        \subcaption{Example images from the ``chainlink fence'' class in ImageNet.}
    \end{subfigure}
    \vfill
    \begin{subfigure}[c]{\linewidth}
        \includegraphics[width=\linewidth]{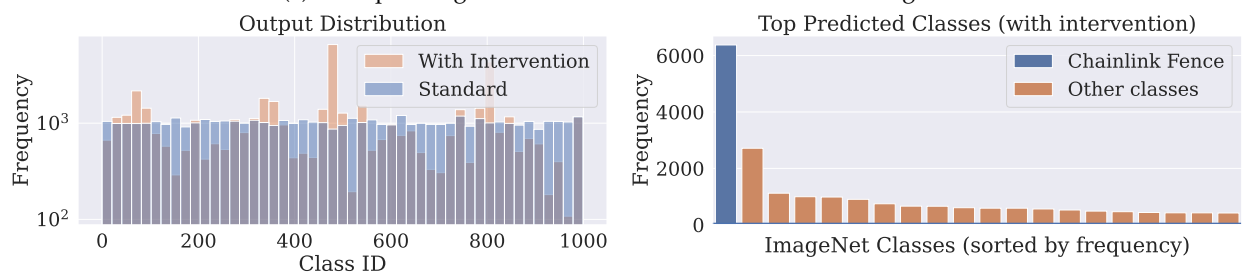}\hfill
        \caption{Shift in ImageNet predicted class distribution after adding a
        ``chainlink fence'' intervention, establishing that the bias holds for the source model.}
    \end{subfigure}\hfill
    \caption{The \textbf{chainlink fence} bias in ImageNet.}
    \label{appfig:IN-chainlink}
\end{figure}

\begin{figure}[!htbp]
    \begin{subfigure}[c]{\linewidth}
        \includegraphics[width=\linewidth]{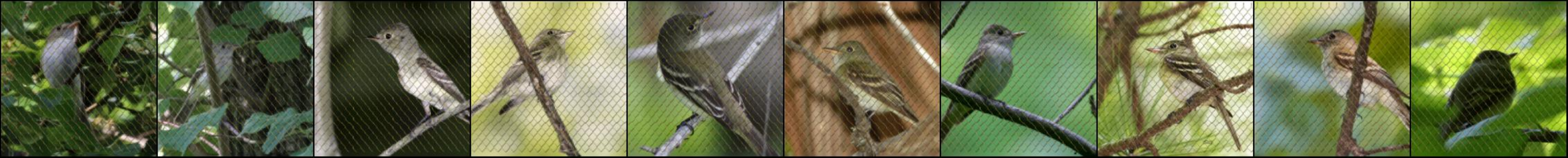}\hfill
        \caption{Example Birdsnap images after applying the chain-link fence intervention.}
    \end{subfigure}\hfill
    \begin{subfigure}[c]{\linewidth}
        \includegraphics[width=\linewidth]{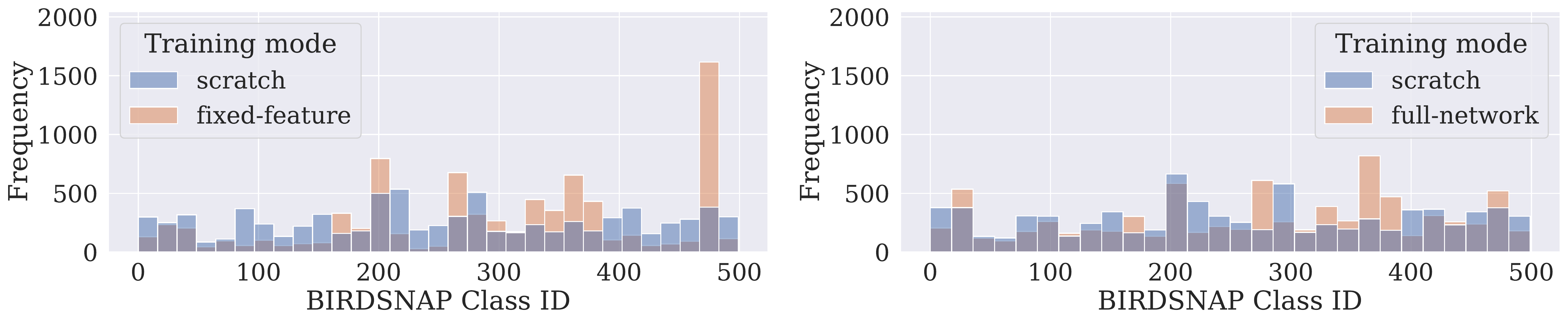}\hfill
        \caption{Output distribution of Birdsnap models with a chainlink fence intervention.}
    \end{subfigure}
    \caption{The \textbf{chainlink fence} bias transfers to \textit{Birdsnap}.}
    \label{appfig:birdsnap-chainlink}
\end{figure}

\begin{figure}[!htbp]
    \begin{subfigure}[c]{\linewidth}
        \includegraphics[width=\linewidth]{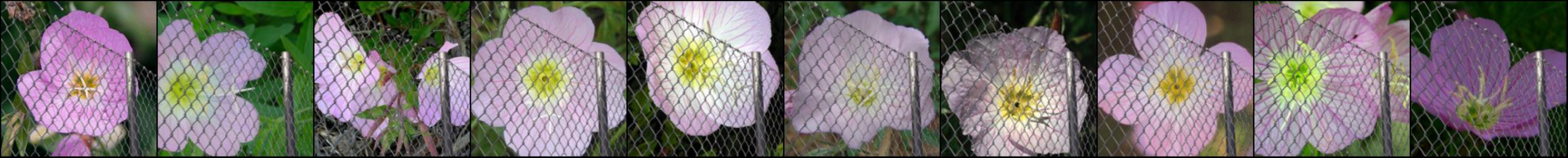}\hfill
        \caption{Example Flowers images after applying the chain-link fence intervention.}
    \end{subfigure}\hfill
    \begin{subfigure}[c]{\linewidth}
        \includegraphics[width=\linewidth]{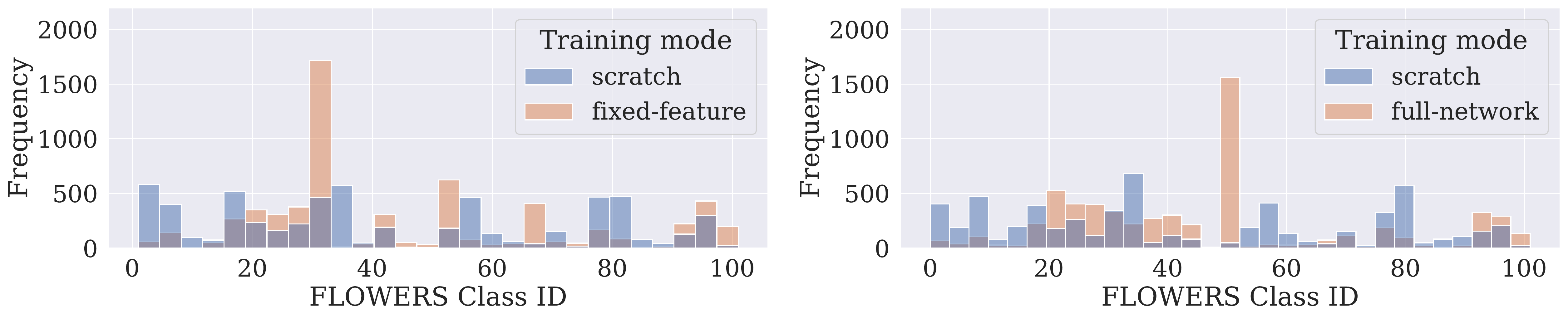}\hfill
        \caption{Output distribution of Flowers models with a chainlink fence intervention.}
    \end{subfigure}
    \caption{The \textbf{chainlink fence} bias transfers to \textit{Flowers}.}
    \label{appfig:flowers-chainlink}
\end{figure}

\begin{figure}[!htbp]
    \begin{subfigure}[c]{\linewidth}
        \includegraphics[width=\linewidth]{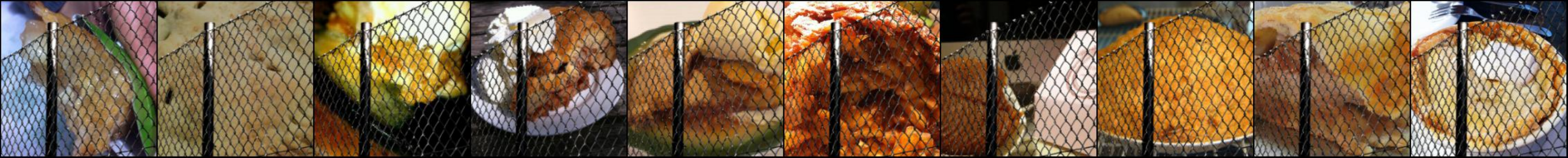}\hfill
        \caption{Example Food images after applying the chain-link fence intervention.}
    \end{subfigure}\hfill
    \begin{subfigure}[c]{\linewidth}
        \includegraphics[width=\linewidth]{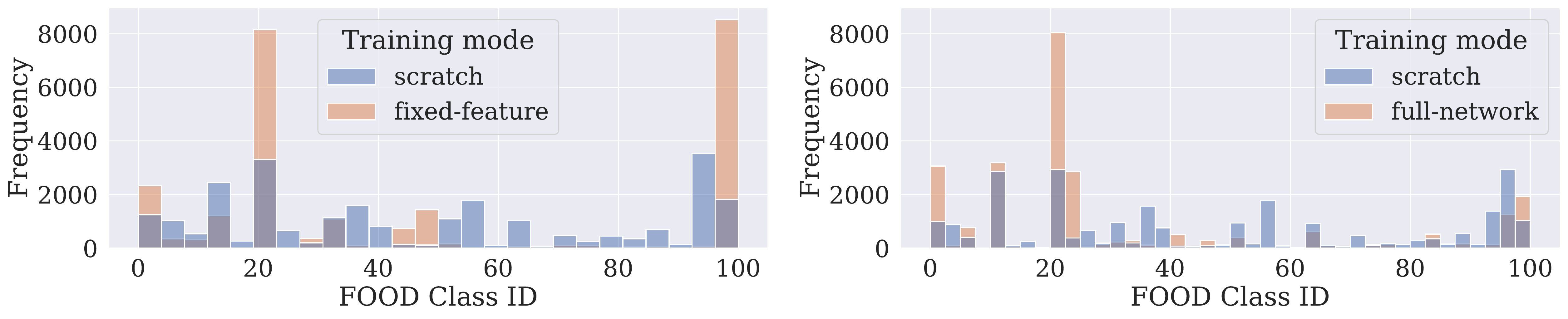}\hfill
        \caption{Output distribution of Food models with a chainlink fence intervention.}
    \end{subfigure}
    \caption{The \textbf{chainlink fence} bias transfers to \textit{Food}.}
    \label{appfig:food-chainlink}
\end{figure}

\begin{figure}[!htbp]
    \begin{subfigure}[c]{\linewidth}
        \includegraphics[width=\linewidth]{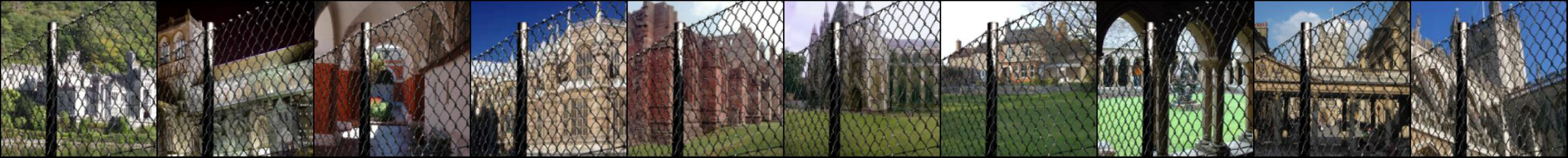}\hfill
        \caption{Example SUN397 images after applying the chain-link fence intervention.}
    \end{subfigure}\hfill
    \begin{subfigure}[c]{\linewidth}
        \includegraphics[width=\linewidth]{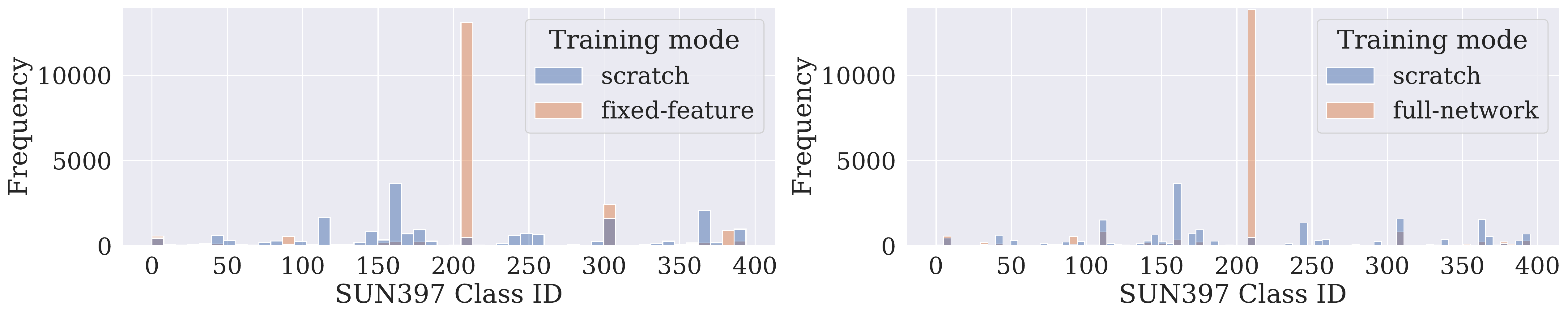}\hfill
        \caption{Output distribution of SUN397 models with a chainlink fence intervention.}
    \end{subfigure}
    \caption{The \textbf{chainlink fence} bias transfers to \textit{SUN397}.}
    \label{appfig:sun397-chainlink}
\end{figure}

In this section we show the results for the ``chainlink fence'' bias transfer. We first demonstrate in Figure~\ref{appfig:IN-chainlink} that the ``chainlink fence'' bias actually exists in ImageNet. Then in Figures~\ref{appfig:birdsnap-chainlink},~\ref{appfig:flowers-chainlink},~\ref{appfig:food-chainlink}, and~\ref{appfig:sun397-chainlink}, we show the output distribution---after applying a chainlink fence intervention---of models trained on various datasets either from scratch, or by transferring from the ImageNet model. The from-scratch models are not affected by the chainlink fence intervention, while the ones learned via transfer have highly skewed output distributions.


\clearpage
\subsection{Hat bias.}
\begin{figure}[!htbp]
    \begin{subfigure}{\linewidth}
        \includegraphics[width=\linewidth,trim={0 0 0 0},clip]{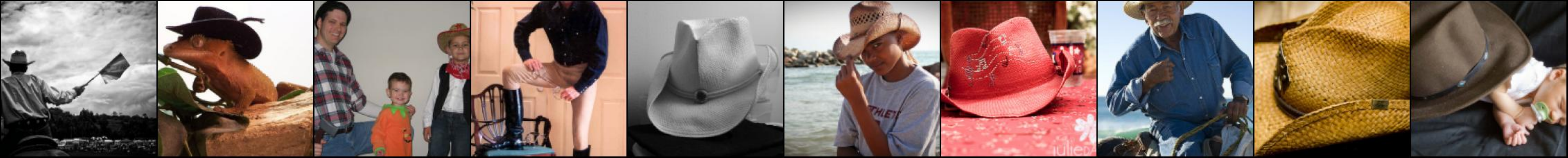}
        \subcaption{ImageNet images from the class ``Cowboy hat''.}
    \end{subfigure}
    \begin{subfigure}[c]{\linewidth}
        \includegraphics[width=\linewidth]{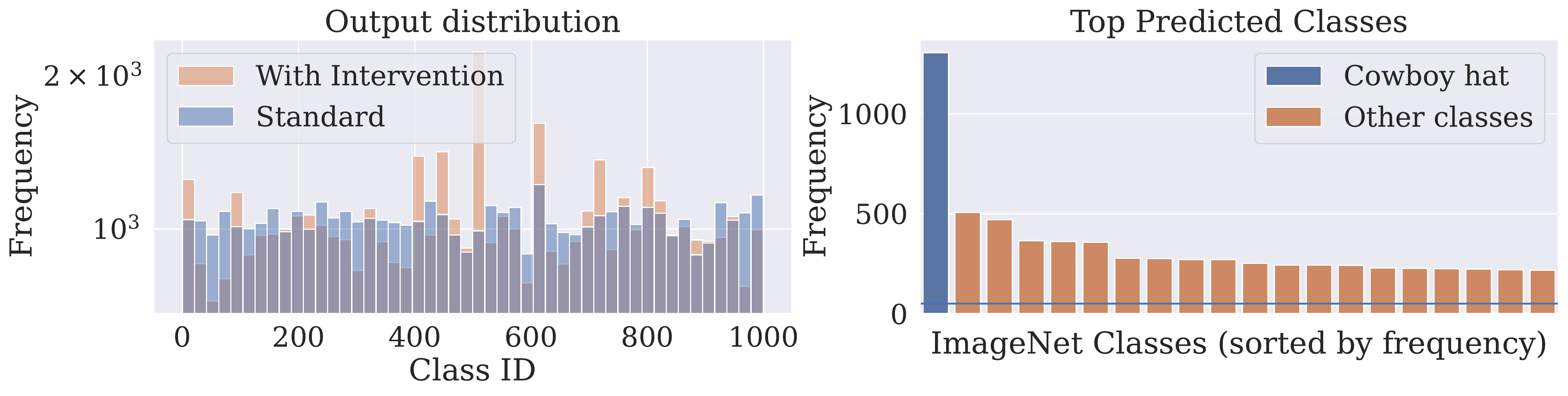}\hfill
        \caption{ImageNet distribution shift after intervention.}
    \end{subfigure}\hfill
    \caption{The \textbf{hat} bias in ImageNet.}
    \label{appfig:IN-cowboyhat}
\end{figure}

\begin{figure}[!htbp]
    \begin{subfigure}[c]{\linewidth}
        \includegraphics[width=\linewidth]{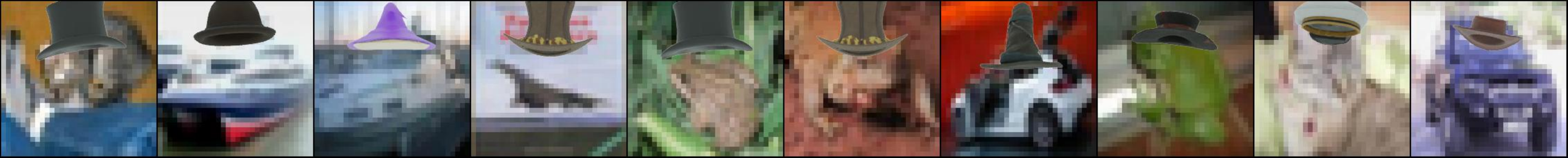}\hfill
        \caption{Example CIFAR-100 images after applying the ``Hat'' intervention.}    
    \end{subfigure}\hfill
    \begin{subfigure}[c]{\linewidth}
        \includegraphics[width=\linewidth]{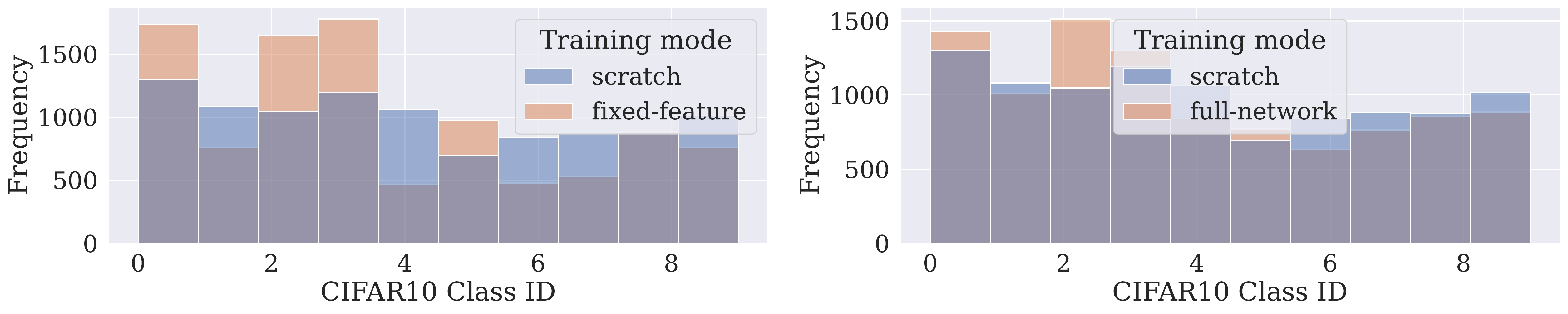}\hfill
        \caption{Output distribution of CIFAR-10 models with the Hat intervention.}
    \end{subfigure}
    \caption{The \textbf{hat} bias transfers to \textit{CIFAR-10}.}
    \label{appfig:cifar10-cowboyhat}
\end{figure}

In this section we show the results for the ``Hat'' bias transfer. We first demonstrate in Figure~\ref{appfig:IN-cowboyhat} that the ``Hat'' bias actually exists in ImageNet (shifts predictions to the ``Cowboy hat'' class). Then in Figure~\ref{appfig:cifar10-cowboyhat}, we show the output distribution---after applying a hat intervention---of models trained on CIFAR-10 either from scratch, or by transferring from the ImageNet model. The from-scratch model is not affected by the hat intervention, while the one learned via transfer have highly skewed output distributions.

\clearpage
\subsection{Tennis ball bias.}
\begin{figure}[!htbp]
    \begin{subfigure}{\linewidth}
        \includegraphics[width=\linewidth,trim={0 0 0 0},clip]{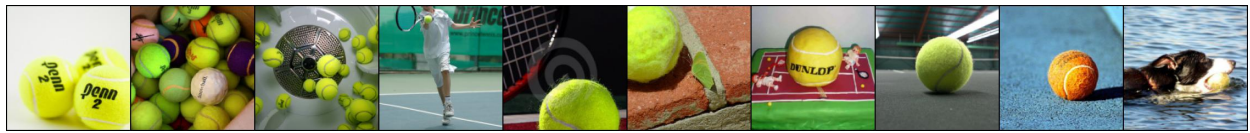}
        \subcaption{ImageNet images from the class ``tennis ball''.}
    \end{subfigure}
    \begin{subfigure}[c]{\linewidth}
        \includegraphics[width=\linewidth]{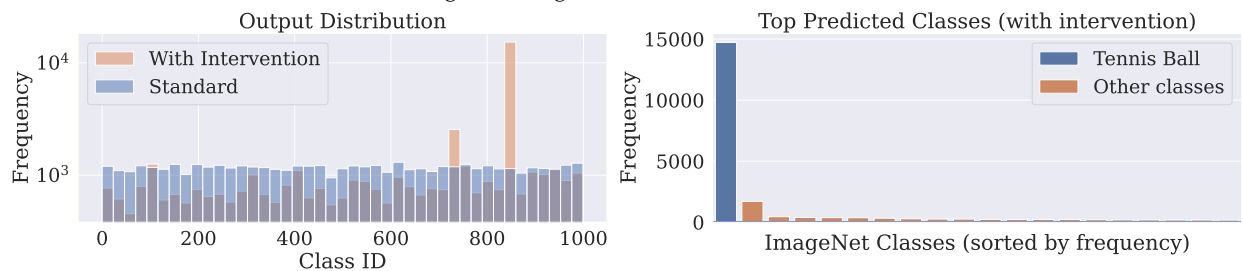}\hfill
        \caption{ImageNet distribution shift after intervention.}
    \end{subfigure}\hfill
    \caption{The \textbf{tennis ball} bias in ImageNet.}
    \label{appfig:IN-tennisball}
\end{figure}

\begin{figure}[!htbp]
    \begin{subfigure}[c]{\linewidth}
        \includegraphics[width=\linewidth]{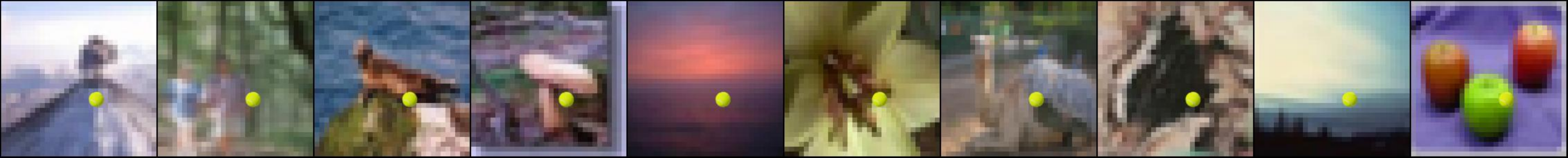}\hfill
        \caption{Example CIFAR-100 images after applying the ``tennis ball'' intervention.}    
    \end{subfigure}\hfill
    \begin{subfigure}[c]{\linewidth}
        \includegraphics[width=\linewidth]{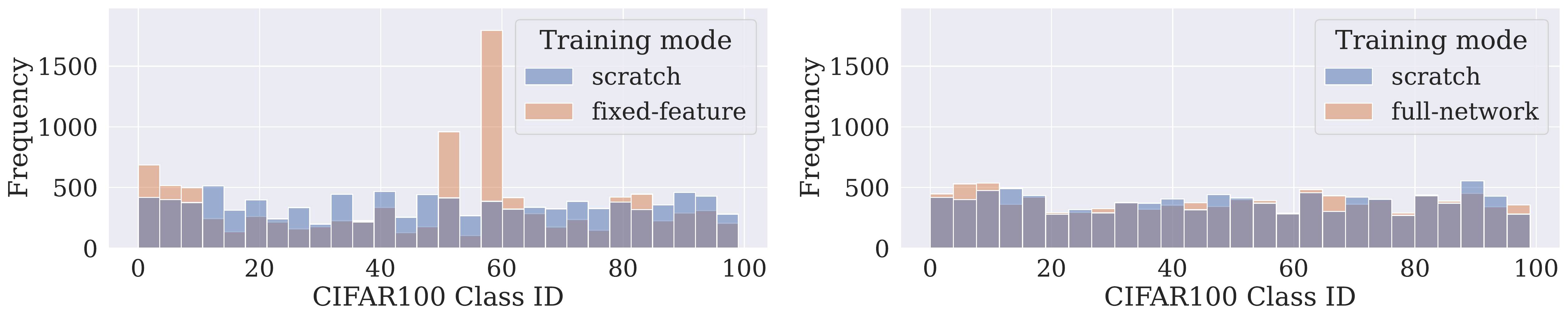}\hfill
        \caption{Output distribution of CIFAR-100 models with the tennis ball intervention.}
    \end{subfigure}
    \caption{The \textbf{tennis ball} bias transfers to \textit{CIFAR-100}.}
    \label{appfig:cifar100-tennisball}
\end{figure}

\begin{figure}[!htbp]
    \begin{subfigure}[c]{\linewidth}
        \includegraphics[width=\linewidth]{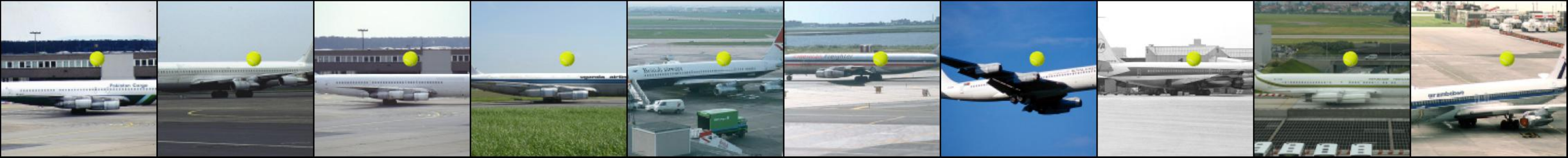}\hfill
        \caption{Example Aircraft images after applying the ``tennis ball'' intervention.}    
    \end{subfigure}\hfill
    \begin{subfigure}[c]{\linewidth}
        \includegraphics[width=\linewidth]{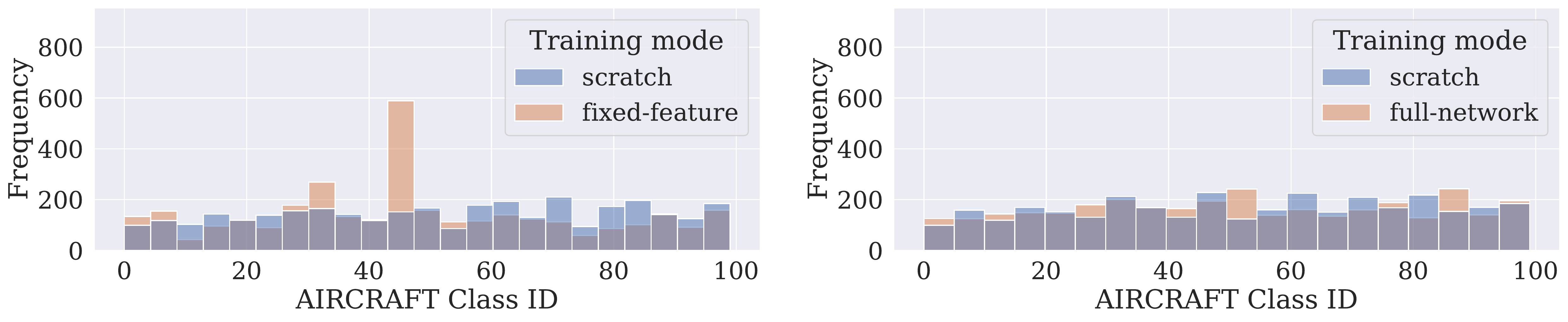}\hfill
        \caption{Output distribution of Aircraft models with the tennis ball intervention.}
    \end{subfigure}
    \caption{The \textbf{tennis ball} bias transfers to \textit{Aircraft}.}
    \label{appfig:aircraft-tennisball}
\end{figure}

\begin{figure}[!htbp]
    \begin{subfigure}[c]{\linewidth}
        \includegraphics[width=\linewidth]{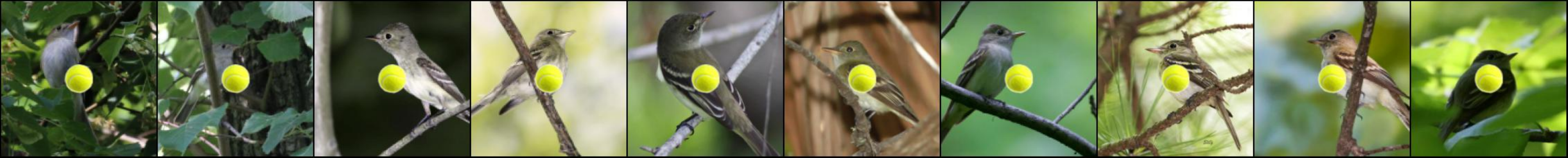}\hfill
        \caption{Example Birdsnap after applying the ``tennis ball'' intervention.}    
    \end{subfigure}\hfill
    \begin{subfigure}[c]{\linewidth}
        \includegraphics[width=\linewidth]{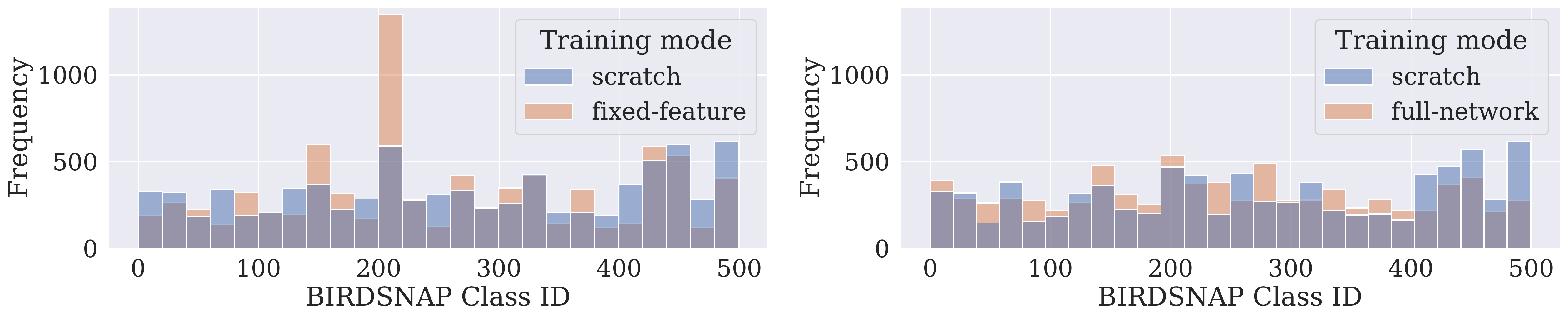}\hfill
        \caption{Output distribution of Birdsnap models with the tennis ball intervention.}
    \end{subfigure}
    \caption{The \textbf{tennis ball} bias transfers to \textit{Birdsnap}.}
    \label{appfig:birdsnap-tennisball}
\end{figure}

\begin{figure}[!htbp]
    \begin{subfigure}[c]{\linewidth}
        \includegraphics[width=\linewidth]{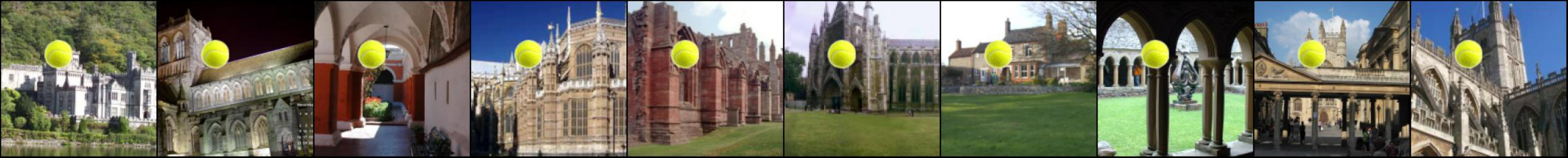}\hfill
        \caption{Example sun397 after applying the ``tennis ball'' intervention.}    
    \end{subfigure}\hfill
    \begin{subfigure}[c]{\linewidth}
        \includegraphics[width=\linewidth]{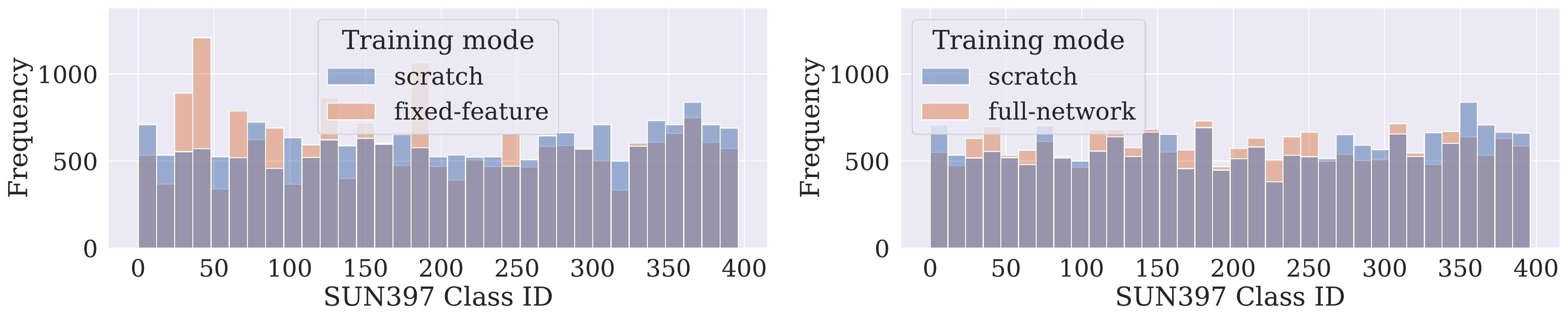}\hfill
        \caption{Output distribution of SUN397 models with the tennis ball intervention.}
    \end{subfigure}
    \caption{The \textbf{tennis ball} bias transfers to \textit{SUN397}.}
    \label{appfig:sun397-tennisball}
\end{figure}

In this section we show the results for the ``tennis ball'' bias transfer. We first demonstrate in Figure~\ref{appfig:IN-tennisball} that the ``tennis ball'' bias actually exists in ImageNet. Then in Figures~\ref{appfig:cifar100-tennisball},~\ref{appfig:aircraft-tennisball},~\ref{appfig:birdsnap-tennisball}, and~\ref{appfig:sun397-tennisball}, we show the output distribution---after applying a tennis ball intervention---of models trained on various datasets either from scratch, or by transferring from the ImageNet model. The from-scratch models are not affected by the tennis ball intervention, while the ones learned via transfer have highly skewed output distributions.


\end{document}